%% file: cvpr.tex
\documentclass[final]{cvpr}

\usepackage{times}
\usepackage{epsfig}
\usepackage{graphicx}
\usepackage{amsmath}
\usepackage{amssymb}
\usepackage{subcaption}
\usepackage{float}
\usepackage{algorithm}
\usepackage{algorithmic}
\usepackage{siunitx}
\usepackage{multirow}
\usepackage{color}
\usepackage[normalem]{ulem}
\usepackage{mmstyle}

\definecolor{gfzhang_color}{RGB}{7, 131, 131}

\usepackage[pagebackref=true,breaklinks=true,colorlinks,bookmarks=false]{hyperref}
\def\cvprPaperID{10955} 
\def\confYear{CVPR 2021}
\begin{document}

\title{VS-Net: Voting with Segmentation for Visual Localization}


\author{
Zhaoyang Huang$^{1,2}$\thanks{Zhaoyang Huang and Han Zhou assert equal contributions.}\\
Xiaowei Zhou$^{1}$

\and
Han Zhou$^{1*}$\\
Hujun Bao$^{1}$
\and
Yijin Li$^{1}$\\
Guofeng Zhang$^{1}$\thanks{Corresponding author: Guofeng Zhang.}\\
\and
Bangbang Yang$^{1}$\\
Hongsheng Li$^{2,3}$
\and
Yan Xu$^{2}$
\and
$^{1}$State Key Lab of CAD\&CG, Zhejiang University\thanks{The authors from Zhejiang University are also affiliated with ZJU-SenseTime Joint Lab of 3D Vision. This work was partially supported by NSF of China(Nos. 61822310 and 61932003), Centre for Perceptual and Interactive Intelligence Limited, the General Research Fund through the Research Grants Council
of Hong Kong under Grants (Nos. 14208417 and 14207319), and CUHK Strategic Fund.}\\
$^{2}$CUHK-SenseTime Joint Laboratory, The Chinese University of Hong Kong \\
$^{3}$School of CST, Xidian University
}
\maketitle

\begin{abstract}
   \input{subsections/abstract}
\end{abstract}

\section{Introduction}
\input{subsections/introduction}

\section{Related Works}
\input{subsections/related_works}

\section{Method}
\input{subsections/method}

\section{Experiments}
\input{subsections/experiment}

\section{Conclusion}
\input{subsections/conclusion}

{\small
\bibliographystyle{ieee_fullname}
\bibliography{egbib}
}

\end{document}

%% file: subsections/abstract.tex
Visual localization is of great importance in robotics and computer vision.
Recently, scene coordinate regression based methods have shown good
performance in visual localization in small static scenes.
However, it still estimates camera poses from many inferior scene coordinates.
To address this problem,
we propose a novel visual localization framework that establishes 2D-to-3D correspondences between the query image and the 3D map with a series of learnable scene-specific landmarks.
In the landmark generation stage, the 3D surfaces of the target scene are over-segmented into mosaic patches whose centers are regarded as the scene-specific landmarks.
To robustly and accurately recover the scene-specific landmarks, we propose the  Voting with Segmentation Network (VS-Net) to segment the pixels into different landmark patches with a segmentation branch and estimate the landmark locations within each patch with a landmark location voting branch.
Since the number of landmarks in a scene may reach up to 5000, training a segmentation network with such a large number of classes is both computation and memory costly for the commonly used cross-entropy loss.
We propose a novel prototype-based triplet loss with hard negative mining, which is able to train semantic segmentation networks with a large number of labels efficiently. Our proposed VS-Net is extensively tested on multiple public benchmarks and can outperform state-of-the-art visual localization methods.
Code and models are available at \href{https://github.com/zju3dv/VS-Net}{https://github.com/zju3dv/VS-Net}.

%% file: subsections/introduction.tex
\input{subsections/introduction/CorrespondenceComparison/CorrespondenceComparison}

Localization~\cite{xu2020selfvoxelo, sarlin2019coarse, qin2018vins} is a pivotal technique in many {real-world} applications, such as Augmented Reality (AR), Virtual Reality (VR), robotics, etc.
With the popularity and low cost of visual cameras,
visual localization has attracted widespread attention from the research community.

Recently, scene coordinate regression based methods~\cite{brachmann2018learning, brachmann2017dsac,li2020hierarchical}, which learn neural networks to predict dense scene coordinates of a query image 
and recover the camera pose through RANSAC-PnP~\cite{fischler1981random}, 
dominate visual localization and achieve state-of-the-art localization accuracy in small static scenes.
Compared with classical feature-based visual localization frameworks~\cite{donoser2014discriminative,li2010location,zeisl2015camera,sarlin2019coarse} relying on identified map points from Structure-from-Motion (SfM) techniques, it only requires to estimate 2D-to-3D correspondences and can be benefited from high-precision sensors.
Although scene coordinates construct dense 2D-3D correspondences, most of them are unable to recover reliable camera poses.
In dynamic environments, there could exist moving objects and varying lighting conditions which raise the outlier ratio and increase the probability of choosing an erroneous pose with RANSAC algorithms.
In addition, even after outlier rejection with RANSAC, there might exist inferior scene coordinates that lead to inaccurate localization. 

In the hope of estimating camera poses more robustly and accurately,
we propose  \textit{Voting with Segmentation Network (VS-Net)} to identify and localize a series of scene-specific landmarks through a \textit{Voting-by-Segmentation} framework. In contrast with scene coordinate regression methods that predict pixel-wise dense 3D scene coordinates, the proposed framework only estimates a small quantity of scene-specific landmarks (or 2D-3D correspondences) that are of much higher accuracy.

Unlike feature-based visual localization methods, where landmarks are {directly extracted} from the {images} according to certain rules, we manually specify a series of scene-specific landmarks from each scene's reconstructed 3D surfaces. The 3D surface of a scene is first uniformly divided into a series of 3D patches, and we define the centers of the 3D patches as the 3D scene-specific landmarks. 
Given a new image obtained from a new viewpoint, we aim to identify the 3D scene-specific landmarks' projections on the 2D image. 
The Voting-by-Segmentation framework with the VS-Net casts the landmark localization problem as a combination of patch-based landmark segmentation coupled with pixel-wise direction voting problem.
Each pixel in the image is first segmented into one of the pre-defined patches (landmarks) and the pixels classified into the $k$th landmark are responsible for estimating the corresponding landmark's 2D location. To achieve the goal, the proposed VS-Net also estimates a 2D directional vector at each pixel location, which is trained to point towards the pixel's corresponding landmark. For a given patch, such predicted directional vectors can be treated as directional votes. With a RANSAC algorithm, for each predicted patch, the accurate 2D landmark location can be accurately estimated.
In contrast to existing scene coordinate regression methods, in our proposed framework, pixels or regions that are poorly segmented with erroneous patch labels and directional votes can be robustly filtered out as those pixels have low voting consistency. Therefore, this strategy ensures that the survived landmarks are of high accuracy and the inferior pixels would not jeopardize the accuracy of camera pose estimation. It results in fewer landmarks with lower outlier ratios and reprojection errors than scene coordinate regression methods~(Fig.~\ref{Fig:scene coordinates and scene-specific landmarks}).

The patch-based landmark segmentation in our VS-Net requires assigning pre-defined patch labels, \ie, landmark IDs, to pixels.
However, the number of patches or landmarks in a scene can reach tens of thousands. Directly adopting the conventional cross-entropy loss for multi-class segmentation requires huge memory and computational costs as the number of parameters in the classification layer increases proportionally to the number of patches.
We propose prototype-based triplet loss to address this problem, 
which avoids computing complete label scores by developing pixel-wise triplet loss with prototypes. 
Moreover, prototype-based triplet loss improves the training efficiency by online mining informative negative prototypes.

In summary, our proposed approach has the following major contributions:
    (1) We propose the novel VS-Net framework that casts the problem of visual localization from scene-specific landmarks as a voting-by-segmentation problem.
    Camera poses estimated from the proposed scene-specific landmarks are shown to be more robust and accurate.
    (2) We propose the prototype-based triplet loss for patch-based landmark segmentation with a large number of classes, which shows competitive segmentation accuracy while saving much computation and memory.
    To our best knowledge, we are the first to address the problem of a large number of classes in image segmentation.
    (3) The VS-Net significantly outperforms 
     previous scene coordinate regression methods and representative SfM-based visual localization methods on both the popular 7Scenes dataset and the Cambridge Landmarks dataset.


%% file: subsections/introduction/CorrespondenceComparison/CorrespondenceComparison.tex
\begin{figure*}[t!]

    \centering
    \resizebox{0.9\linewidth}{!}{
    \begin{subfigure}[b]{0.3\linewidth}
        \includegraphics[width=\linewidth]{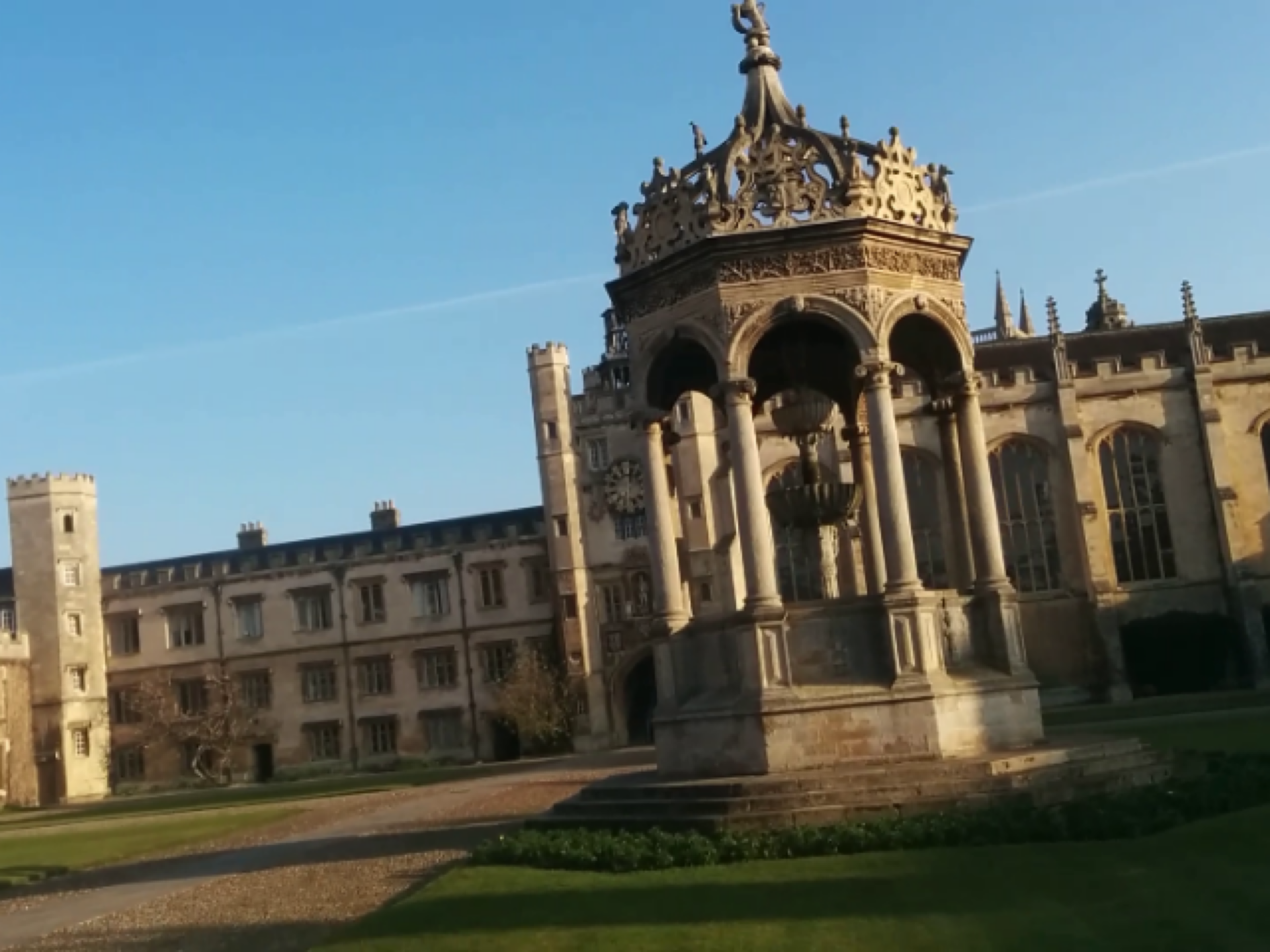}
        \caption{\small Query Image}
    \end{subfigure}
    \begin{subfigure}[b]{0.3\linewidth}
        \includegraphics[width=\linewidth]{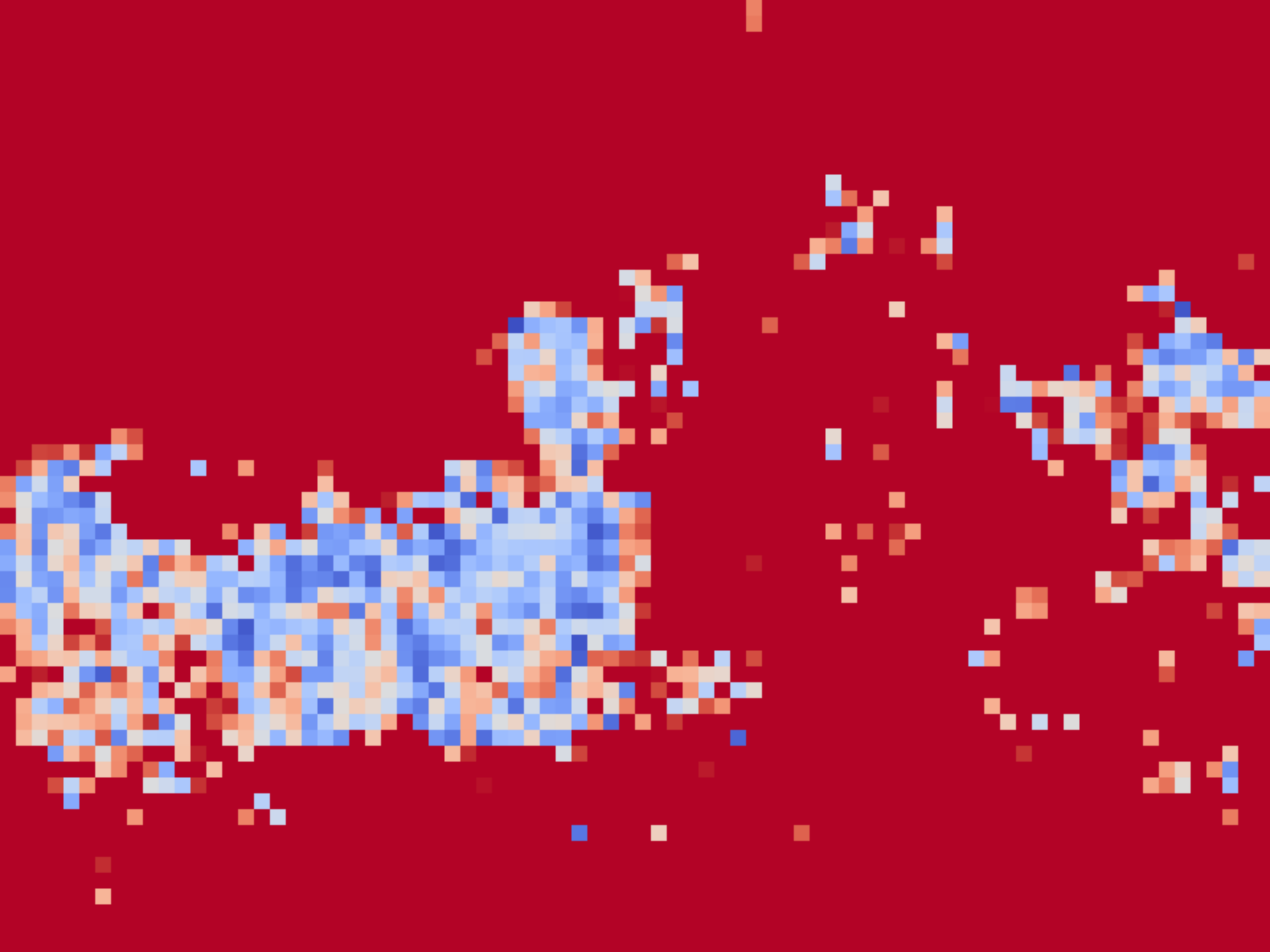}
        \caption{\small Errors of Scene Coordinates}
    \end{subfigure}
    \begin{subfigure}[b]{0.3\linewidth}
    \includegraphics[width=\linewidth]{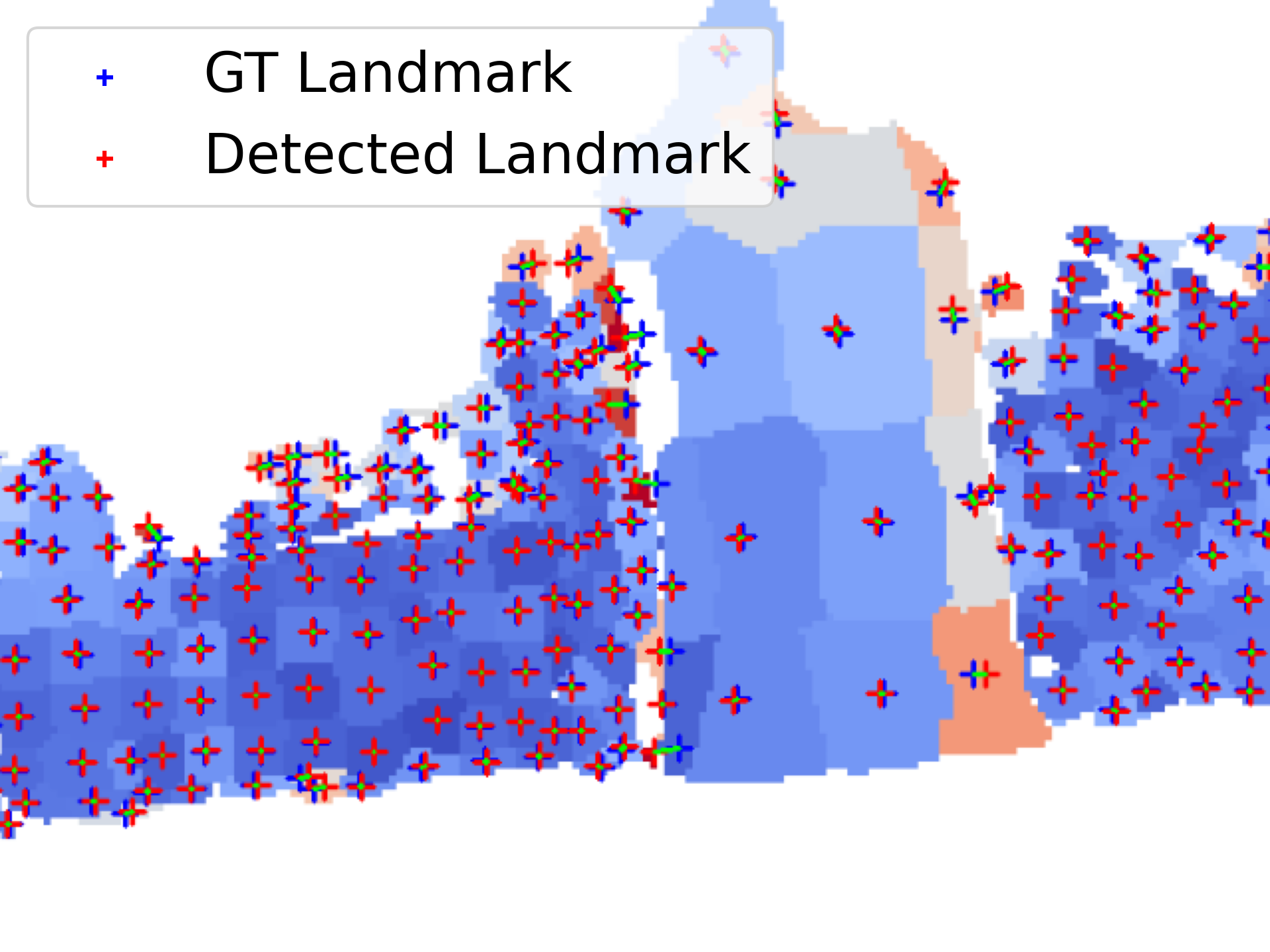}        
    \caption{\small Errors of Scene-Specific Landmarks}
    \end{subfigure}
    \begin{subfigure}[b]{0.065\linewidth}
    \includegraphics[width=\linewidth, keepaspectratio]{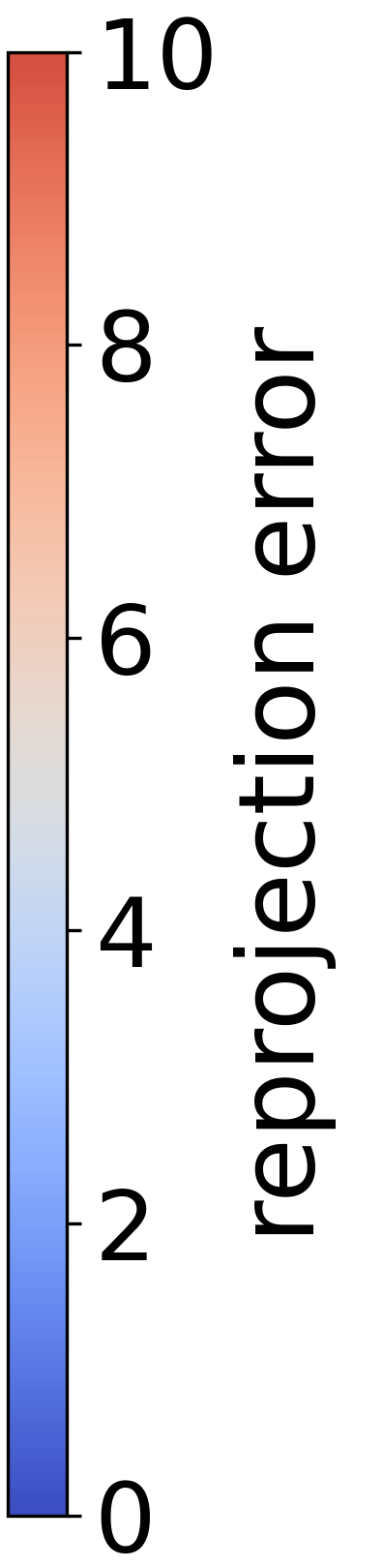} 
    \end{subfigure}
    }
    \caption{
    Reprojection errors of 2D-to-3D correspondences of scene coordinates and scene-specific landmarks.
    (a) The query image.
    (b) The reprojection errors of dense scene coordinates predicted by the regression-only network~\cite{li2020hierarchical}.
    (c) The reprojection errors of scene-specific landmarks and their surrounding patches by the proposed method. 
    Pixels belonging to the same landmark are painted with the same color representing the landmark' reprojection error.
    The white pixels in (c) are filetered by our voting-by-segmentation algorithm.
    }
    \label{Fig:scene coordinates and scene-specific landmarks}
\end{figure*}

%% file: subsections/related_works.tex
\noindent {\bf Visual Localization.}
Visual localization aims at estimating 6-DoF camera pose in the map built beforehand for a query image.
{Traditional visual localization frameworks}~\cite{arth2009wide,donoser2014discriminative,li2010location,zeisl2015camera,sarlin2019coarse,camposeco2019hybrid} build a map by SfM techniques~\cite{zhang2010efficient,agarwal2011building,zhu2017parallel,schonberger2016structure,wu2011visualsfm} with general feature detectors and descriptors \cite{ lowe2004distinctive, bay2006surf, rublee2011orb, ono2018lf, detone2018superpoint, DBLP:conf/cvpr/DusmanuRPPSTS19, liu2019gift, revaud2019r2d2}.
Given a query image, they extract the same 2D features and match them to the 3D features in the map via descriptors.
The capability of the feature detector and the feature descriptor is of great importance in this framework because it affects both the map quality and the establishment of the 2D-3D correspondences in a query image, which determines the localization accuracy. 
Many feature detectors and descriptors have been proposed, such as handcrafted features~\cite{ lowe2004distinctive, bay2006surf, morel2009asift, rublee2011orb, imperoli2016active} and learned features~\cite{ono2018lf, detone2018superpoint, DBLP:conf/cvpr/DusmanuRPPSTS19, revaud2019r2d2, ge2020self, wang2020learning}.
In SfM-based visual localization systems, the 3D feature points are reconstructed with triangulation according to multiple associated observations.
They are always messy in that an ideal 3D point may be represented by different feature points that are not matched and merged because of large viewpoint or scale change, which may impact the following localization. 

\input{subsections/method/Framework/Framework}

With the development of deep learning, training a scene-specific neural network to encode the map and localize an image from it becomes an alternative visual localization approach.
Neural pose regression~\cite{kendall2015posenet,kendall2017geometric,brahmbhatt2018geometry, huang2019prior, wang2020atloc, xue2020learning} learns to directly predict parameters of a camera pose from an image, which are not competitive with other visual localization frameworks in accuracy.
Another method is to predict scene coordinates~\cite{brachmann2017dsac,brachmann2018learning,shotton2013scene,valentin2015exploiting, budvytis2019large, brachmann2019expert, li2020hierarchical, zhou2020kfnet} as an intermediate representation and estimate the camera pose through a RANSAC-PnP~\cite{fischler1981random, shotton2013scene} algorithm, which achieves state-of-the-art localization performance in small and medium scenes.
Recently, many works extend this pipeline for better localization accuracy.
Brachmann~\etal~\cite{brachmann2017dsac, brachmann2018learning} learn scene coordinate regression with differential RANSAC.
Li~\etal~\cite{li2020hierarchical} hierarchically predicts scene coordinates.
Zhou~\etal~\cite{zhou2020kfnet} improves the regression by using temporal information.
Weinzaepfel~\etal~\cite{weinzaepfel2019visual} propose to localize from Objects-of-Interest, which is so coarse-grained and requires annotations.

\noindent {\bf Keypoint-based Object Pose Estimation.}
Keypoint is widely utilized as an intermediate representation in object pose estimation~\cite{peng2019pvnet, he2020pvn3d, oberweger2018making, pavlakos20176, song2020hybridpose}.
Many of them showed that keypoint-based pose estimation outperforms direct pose regression and object-customized keypoints are better than general features.
Inspired by these works, we propose to learn to find scene-specific landmarks for visual localization.
Recently, PVNet~\cite{peng2019pvnet} significantly improves robustness and accuracy of object pose estimation by detecting keypoints with pixel-wise votes,
inspired by which, we propose to detect scene-specific landmarks with pixel-wise votes.

\noindent {\bf Semantic Segmentation and Large-scale Classification.}
Semantic segmentation~\cite{long2015fully, ronneberger2015u, DBLP:journals/corr/YuK15, chen2017deeplab, zhong2020squeeze}, which predicts pixel-wise labels according to a set of semantic categories, is a long-standing topic in computer vision and has been widely discussed in the past decades.
Unfortunately, pixel-wise cross-entropy loss adopted by previous methods devours a lot of memory and computation when the number of categories is large. Furthermore, the classifier matrix can not be learned effectively due to the large variance of gradients~\cite{xiao2017joint}.
Large-scale classification is also encountered in many other tasks, such as person re-identification~\cite{xiao2017joint, zheng2019pose, zhong2020learning, lin2020unsupervised}, face recognition~\cite{schroff2015facenet, wu2017sampling}, \etc.
Online instance matching~(OIM)~\cite{xiao2017joint} loss and proxy-based metric learning~\cite{aziere2019ensemble, movshovitz2017no, qian2019softtriple} share a similar idea that maintains a memory bank containing a feature prototype for each label.
However, the number of pixels in an image is quite large, and they still compute scores between each data point and each label, which runs into the same situation of cross-entropy loss.
To tackle this issue,
we propose the prototype-based triplet loss that simultaneously maintains a prototype for each class and a network to predict class labels by imposing pixel-wise triplet loss on prototypes.

%% file: subsections/method/Framework/Framework.tex
\begin{figure*}[t!]
    \centering
    \includegraphics[width=0.9\linewidth, trim={0 105mm 0 0}, clip]{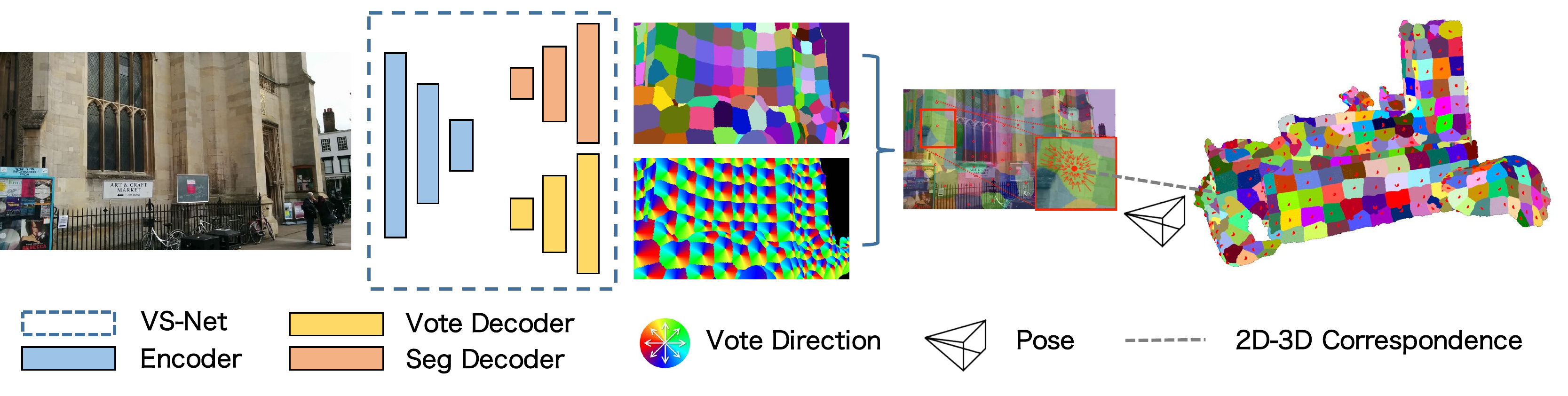}
    \caption{Visual localization by VS-Net.
    There are two decoder branches respectively predicting a landmark segmentation map and a pixel-wise voting map, from which we can detect the location and labels of landmarks.
    After establishing 2D-3D correspondences according to the landmark labels, we can estimate the 6-DoF camera pose with a standard RANSAC-PnP.
    }
    \label{Fig:framework}
\end{figure*}

%% file: subsections/method.tex
State-of-the-art visual localization methods for small-scale scenes
are dominated by scene coordinate regression based methods \cite{brachmann2017dsac,brachmann2018learning} that establish dense 2D-to-3D correspondences (scene coordinates) between each pixel in an input query image and the 3D surface points of a scene.
However, a large portion of predicted scene coordinates shows high re-projection errors, which increase the chance of localization failure and deteriorate the localization accuracy of the follow-up RANSAC-PnP algorithms.
To tackle the issues, we propose the Voting with Segmentation Network (VS-Net) to identify a series of scene-specific landmarks (Fig.~\ref{Fig:framework}) and establish their correspondences to the 3D map for achieving accurate localization.
The scene-specific landmarks are {sparsely and directly defined} from a scene's 3D surfaces.
Given different viewpoints of the training images, we can project the scene-specific landmarks and their surrounding surface patches to the image planes to identify their corresponding pixels in the images.
In this way, we obtain the pixels of the surrounding patches of each landmark in the multiple training images.
The problem of localizing the scene-specific landmarks from the images can be cast as 2D patch-based landmark segmentation and pixel-wise landmark location voting.

During the training phase, for all pixels corresponding to the same surrounding surface patch of a landmark, their outputs are required to predict the same segmentation label (landmark ID) via patch-based landmark segmentation of the proposed VS-Net. 
Another landmark location branch is introduced to make each pixel responsible for estimating the 2D location of its corresponding landmark by outputting the directional vectors pointing towards the landmark's 2D projection.

For inference, given a new input image, we obtain the landmark segmentation map and the landmark location voting map from the VS-Net.
The 2D-to-3D landmark correspondences can then be established based on the landmark segmentation and location voting maps.
Unlike outlier 2D-to-3D correspondences from scene coordinate regression methods that can only be rejected by RANSAC {PnP} algorithms, landmarks from our proposed approach that do not have high enough voting confidence would be directly dropped, which prevent estimating camera poses from poorly localized landmarks (Fig.~\ref{Fig:scene coordinates and scene-specific landmarks}).
{Furthermore, the correspondences built upon scene coordinate methods can be easily influenced by unstable predictions, while minor disturbed votes do not deteriorate the accuracy of the voted landmark locations in our method because they would be filtered by the within-patch RANSAC intersection algorithm.}




\subsection{Creation of Scene-specific 3D Landmarks}
\label{sec: 3.1}

Given each scene for visual localization, we can obtain the scene's 3D surfaces from existing 3D reconstruction algorithms, such as multiview stereo~\cite{schoenberger2016mvs}, Kinect fusion~\cite{newcombe2011kinectfusion}, \etc.
The proposed scene-specific 3D landmarks are created based on the reconstructed 3D surfaces. We partition the 3D surfaces into a series of 3D patches with the 3D over-segmentation algorithm, Supervoxel~\cite{Papon13CVPR}.
The center points of the $n$ over-segmented 3D patches $\{{\bf q}_1, \dots, {\bf q}_n\} \in \mathbb{R}^3$ are chosen as the scene-specific landmarks for localization.
As Supervoxel produces patches of similar sizes, the generated landmarks are mostly uniformly scattered on the 3D surfaces, 
which can provide enough landmarks from different viewpoints and therefore benefit localization robustness.

Given the training images {along with camera poses} of a scene, the 3D scene-specific landmarks $\vq_1, \dots, \vq_n$, and their associated 3D patches can be projected to the 2D images. 
For each image, we can generate a landmark segmentation map {$\mS \in \mathbb{Z}^{H\times W}$} and a landmark location voting map $\vd \in \mathbb{R}^{H \times W \times 2}$.
For patch-based landmark segmentation, each pixel $i$ with coordinate ${\bf p}_i = (u_i, v_i)$ is assigned the landmark label (ID) $j$ determined by the projection of the 3D patches.
{If a pixel corresponds to some region that is not covered by the projected surfaces, such as the sky or distant objects, a background label 0 is assigned to it to represent that this pixel is noneffective for visual localization.}

For landmark location voting, we first compute a landmark $\vq_j$'s projected 2D location $\vl_j = \mathcal{P}(\vq_j, \mK, \mC) \in \mathbb{R}^2$ by projecting the 3D landmark according to the camera intrinsic matrix $\mK$ and the camera pose parameters $\mC$.
Each pixel {$i$} belonging to landmark $j$'s patch is responsible for predicting the 2D directional vector $\vd_i \in \mathbb{R}^2$ pointing towards the $j$'s 2D projection, \ie
\begin{align}
\vd_i=(\vl_j-\vp_i)/||\vl_j-\vp_i||_2,
\label{eq:directional}
\end{align}
where $\vd_i$ is a normalized 2D vector denoting the landmark $j$'s direction.


After defining the ground-truth landmark segmentation and location voting maps, $\mS$ and $\vd$, we can supervise the proposed VS-Net to predict the two maps. After training, VS-Net can predict the two maps for a query image, from which we can establish accurate 2D-to-3D correspondences for achieving robust visual localization.


\begin{figure}
    \centering
    \includegraphics[width=1.0\linewidth, trim={5mm 88mm 140mm 5mm}, clip]{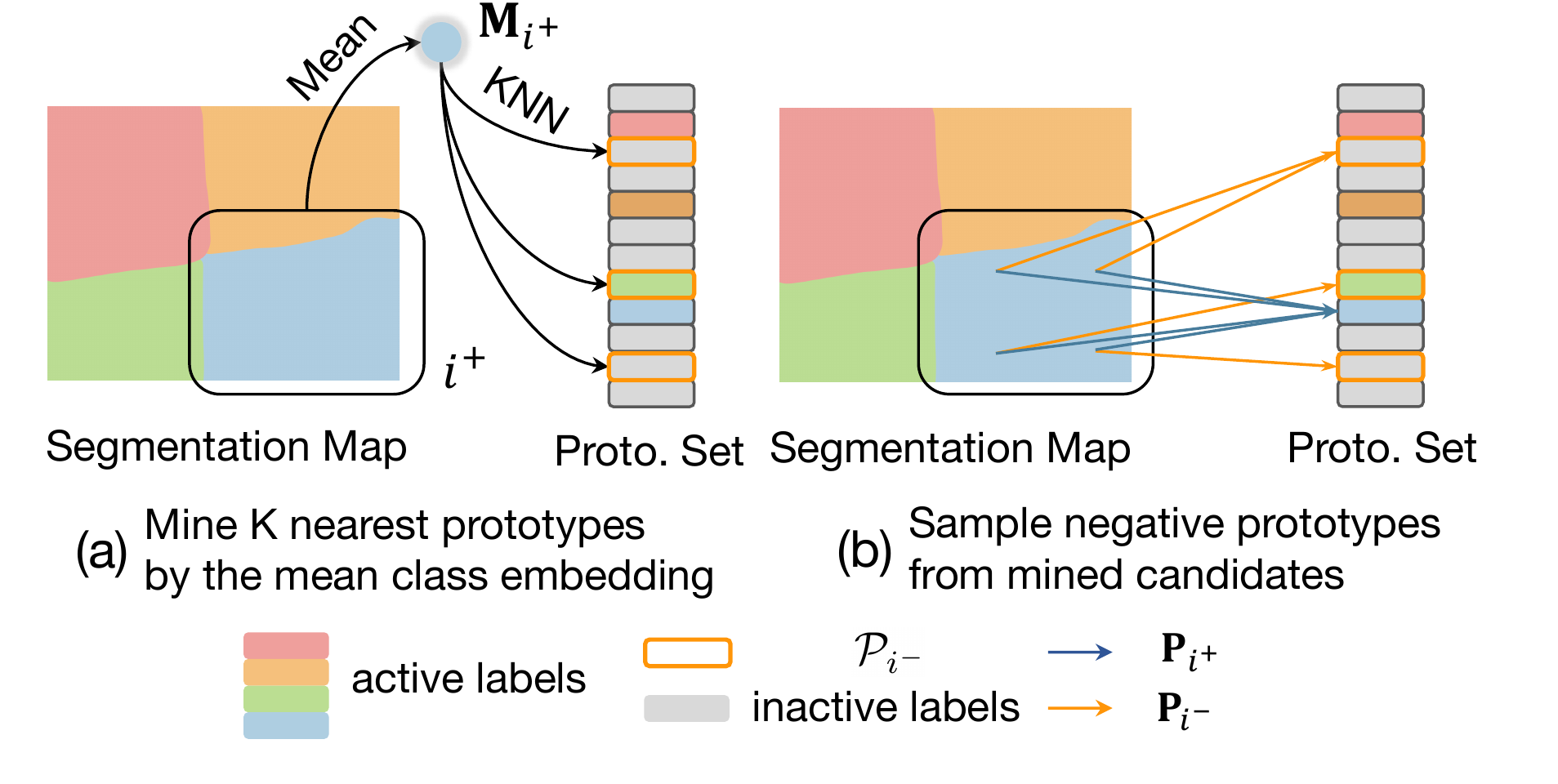}
    \caption{Prototype-based triplet loss.
    (a) Given a pixel $i$ with its mean class embeddings~$\mM_{i^+}$ calculated with pixels belonging to the same label $i^+$, we find its $k$ negative nearest neighbors in the prototype set.
    (b) We formulate pixel-wise prototype-based triplets where positive prototype is $\mathbf{P}'_{i^+}$ and the negative prototype is sampled from the mined $K$ neighbors ${\cal P}_{i^{-}}$.
    }
    \label{Fig:proxy-m2 loss}
\end{figure}

\subsection{Voting-with-Segmentation Network with Prototype-based Triplet Loss}
\label{sec: 3.2}

The proposed Voting-with-Segmentation Network (VS-Net) consists of an image encoder to encode the image into visual feature maps, a segmentation decoder to predict patch-based landmark segmentation map, and a voting decoder for generating the landmark voting map. The two maps are responsible for estimating landmarks' 2D locations as detailed below.
In contrast with scene coordinate regression that relies on a neural network with a small receptive field to avoid overfitting, as discussed by Brachmann~\etal~\cite{li2020hierarchical},
the prediction of the pixel-wise landmark labels and landmark directional votes can benefit from contextual information.
We use DeepLabv3~\cite{DBLP:journals/corr/ChenPSA17}, which enlarges the receptive field with atrous spatial pyramid pooling, as the backbone of our VS-Net.

\noindent {\bf Patch-based landmark segmentation with prototype-based triplet loss.}
{Conventional semantic segmentation tasks generally adopt the cross-entropy loss to supervise complete categorical confidence vectors of all predicted pixels.}
However, our landmark segmentation requires \sout{to} output segmentation maps with a large number of classes (landmarks) to effectively model each scene. A common scene in the 7Scenes dataset can consist of up to $5000$ landmarks. Simply supervising a $5000$-class segmentation map of size $640\times480$ with the cross-entropy loss requires $34.3$ GFLOPS and $5.7$ GB memory, which can easily drain computational resources of even modern GPUs.

To address this issue, we propose a novel prototype-based triplet segmentation loss with online hard negative sampling to supervise semantic segmentation with a large number of classes. It maintains and updates a set $\mP$ of learnable class prototype embeddings, each of which is responsible for a semantic class, and $\mP_j$ denotes the $j$th class's embedding. Intuitively, pixel embeddings of the $j$th class should be close to $\mP_j$ and be far away from other classes' prototypes. 
Our proposed loss is designed based on the triplet loss with an online negative sampling scheme.

Specifically, given a pixel-wise embedding map $\mE$ output by the segmentation branch of VS-Net and the class prototype set $\mP$, the individual embeddings are first $L2$ normalized and are then optimized to minimize the following prototype-based triplet loss for each pixel $i$'s embedding $\mE_i$,
\begin{align}
    {\cal L}_{\rm seg} = \sum_{{\rm all~} i} \max(0, m + {\rm sim}(\mE_i, \mP_{i^-}) - {\rm sim}(\mE_i, \mP_{i^+})),
 \label{lssm loss}
\end{align}
where ${\rm sim}(a,b) = \frac{\boldsymbol{a}^T \boldsymbol{b}}{\| \boldsymbol{b} \| \cdot \| \boldsymbol{b} \|}$ measures the cosine similarity between a pixel embedding and a class prototype embedding, $m$ represents the margin of the triplet loss, $\mP_{i^+}$ denotes the ground-truth (positive) class's prototype embedding corresponding to pixel $i$, and $\mP_{i^-}$ denotes a sampled non-corresponding (negative) class prototype embedding of $i$ (to be discussed below).

For each pixel $i$, how to determine its negative-class prototype embedding $\mP_{i^{-}}$ in the above prototype-based triplet loss has crucial impacts on the final performance and randomly sampling negative classes would make the training over-simplified. 
{Given an input image, we observe that the number of active landmarks (\ie at least one pixel in the image belonging to the landmarks) is limited.}
{In addition, pixels belonging to the same patch of a landmark are spatially close to each other and would share similar hard negative prototypes because they have similar embeddings.
}
{We, therefore, propose to mine representative negative classes for each active landmark, and each pixel randomly samples a negative class from the mined class set to form representative triplets.}

Specifically, given a pixel $i$ with an active landmark (class) index $i^+$, we first retrieve all pixel embeddings corresponding to the landmark $i^+$ in the input image and take their average to obtain the landmark's mean class embedding $\mM_{i^+}$ from the current image. The mean class embedding is then used to retrieve the $k$-nearest-neighbor negative prototypes ${\cal P}_{i^{-}}$ from the prototype embedding set.
Such $k$NN negative prototypes can be considered as hard negative classes. The pixel $i$'s single negative prototype embedding $\mP_{i^{-}}$ to be used in the triplet loss (Eq. \eqref{lssm loss}) is uniformly sampled from the $k$NN negative prototype set (Fig.~\ref{Fig:proxy-m2 loss}).

The proposed prototype-based triplet loss is much more efficient than the conventional cross-entropy loss when used for supervising semantic segmentation {as it only computes complete class scores for active landmarks rather than for all pixels}.
With an input image of size $640\times 480$, the conventional cross-entropy loss costs 36.9 GFLOPS and 5.7GB memory. In contrast, if there are 100 active labels in an image, our proposed prototype-based triplet loss costs only 26.7 MFLOPS and $3.08$ MB memory, where the $k$NN hard negative search costs $12.0$ MFLOPS and $1.91$ MB memory, and the triplet loss itself only costs $14.7$ MFLOPS and $1.17$ MB memory~(Tab.~\ref{tab:complexity analysis}).
OIM loss~\cite{xiao2017joint} is a popular loss for supervising large-scale classification problems. However, for each sample, it still needs to compute scores of belonging to all classes. As each image has a large number of pixels, it is still impractical to adopt the OIM loss in semantic segmentation.

{
\noindent {\bf Pixel-wise voting for landmark location.}
Given the segmentation map $\mS$ generated from the above-introduced segmentation decoder, each pixel $i$ in the input image is either assigned a landmark label $\mS_i$ or a noneffective label $0$ denoting too distant objects or regions (\eg, sky).
We introduce another voting decoder for determining landmarks' projected 2D locations in the given image. The decoder outputs a directional voting map $\vd$, where each pixel $i$ outputs a 2D directional vector $\vd_i$ (Eq. \eqref{eq:directional}) pointing towards its corresponding landmark's 2D location (according to $\mS_i$). The voting decoder is supervised with the following loss,
}
\begin{equation}
{\cal L}_{\rm vote}(i) = \sum_{{\rm all}~i} \mathbf{1}(\mS_{i} \neq 0)||\hat{\vd}_{i}-\vd_{i}||_1,
\label{eq:voting loss}
\end{equation}
{where $\mathbf{1}$ denotes the indicator function, and $\vd_i$ and $\hat{\vd}_i$ are ground-truth and predicted voting directions of pixel $i$.}

\noindent {\bf Overall loss function.}
The overall loss ${\cal L}_{\rm overall}$ is the combination of the patch-based landmark segmentation loss and landmark direction voting loss,
\begin{equation}
{\cal L}_{\rm overall} = {\cal L}_{\rm seg}(i) + \lambda {\cal L}_{\rm vote}(i),
\label{eq:all loss}
\end{equation}
where $\lambda$ weights the contributions of the loss terms.

\input{subsections/method/PointDetection/PointDetection}

\noindent {\bf Localization with landmark segmentation and voting maps.} In the localization stage,
pixels that are predicted to have the same landmark label in the landmark segmentation map are grouped together and we estimate its corresponding landmark location by computing the intersection of the landmark directional votes from the predicted voting map, which is dubbed the voting-by-segmentation algorithm.

Particularly,
given the segmentation map, we first filter out landmark patches whose sizes are smaller than a threshold $T_s$ because too small landmark segments are generally unstable.
The initial estimation of the 2D location $\hat{\bf l}_j$ of the landmark $j$ is computed from RANSAC with a vote intersection model~\cite{peng2019pvnet}, which generates multiple landmark location hypotheses by computing intersections of two randomly sampled directional votes and choosing the hypothesis having the most inlier votes.
Then, the locations are further refined by an iterative EM-like algorithm.
In the E-step, we collect inlier directional votes for the landmark $j$ from the surrounding circular region of the current $\hat{\bf l}_j^{(t)}$.
In the M-step, we adopt the least-square method introduced by Antonio~\etal~\cite{antonio1992faster} to compute the updated landmark location $\hat{\bf l}_j^{(t+1)}$ from the votes in the circular region.
During the iterations, a voted landmark not supported by enough directional votes, indicating low voting consistency, would be dropped.

There are inevitable some disturbed pixels and some disordered regions caused by environmental noise or unfaithful surfaces.
{
As shown in Fig.~\ref{Fig:point detection},
the landmarks generated by our voting-by-segmentation algorithm achieve high accuracy and robustness against these distracting factors because we can accurately detect landmark locations by filtering disturbed pixel votes (pointed by the red arrow) and further reject unstable regions (pointed by the green arrow) in advance by checking the voting consistency.
In contrast,
cluttered SIFT features can easily result in erroneous matches, and the detected locations are easily disturbed on locally unstable regions,~\eg trees.
}
Finally, all the estimated 2D landmarks in the query image naturally associate with the 3D landmarks in the scene, and the camera pose can be reliably estimated {with standard RANSAC-PnP algorithm}.

\begin{table}[t]
    \centering
    \resizebox{0.85\linewidth}{!}{
    \begin{tabular}{c|c|cc}
    \hline
    & cross entropy & \multicolumn{2}{c}{proto. triplet} \\
    \hline 
    & Total & $k$NN & Triplet\\
    Computation & 36.9 GF & 12.0 MF & 14.7 MF\\
    Memory & 5.7 GB & 1.91 MB & 1.17 MB\\
    \hline
    \end{tabular}
    }
    \caption{Computation and memory cost comparison.
    GF and MF denotes GFLOPS and MFLOPS.
    }
    \vspace{-5mm}
    \label{tab:complexity analysis}
\end{table}

\input{subsections/experiment/ComparisonTable}
\input{subsections/experiment/PoseAccuracyFigure/PoseAccuracyFigure}
\input{subsections/experiment/HardCase/HardCase}

%% file: subsections/method/PointDetection/PointDetection.tex
\begin{figure}[!tb]
    \centering
    \resizebox{0.9\linewidth}{!}{
    \begin{subfigure}[b]{0.49\linewidth}
        \includegraphics[width=\linewidth]{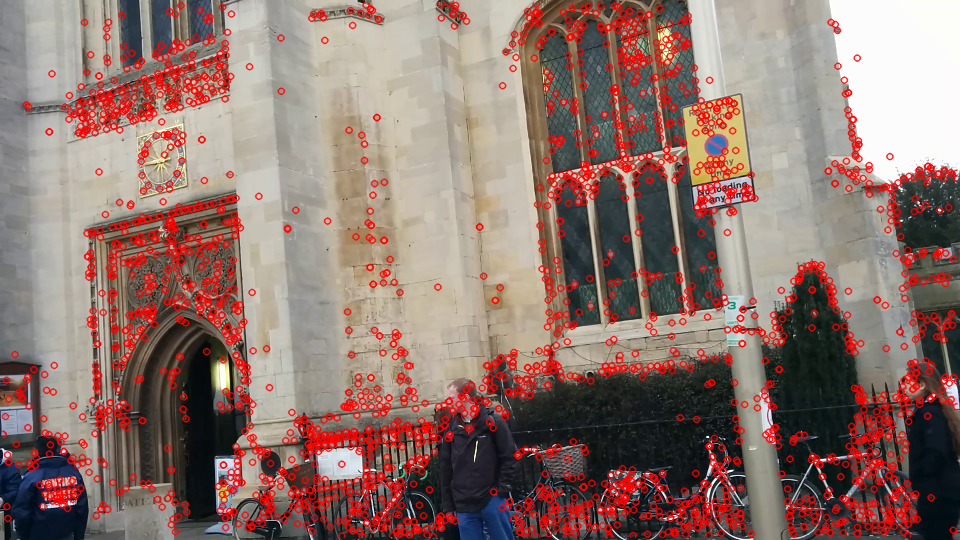}
        \caption{\small SIFT features}
    \end{subfigure}
    \begin{subfigure}[b]{0.49\linewidth}
        \includegraphics[width=\linewidth, trim={0mm 130mm 230mm 0mm}]{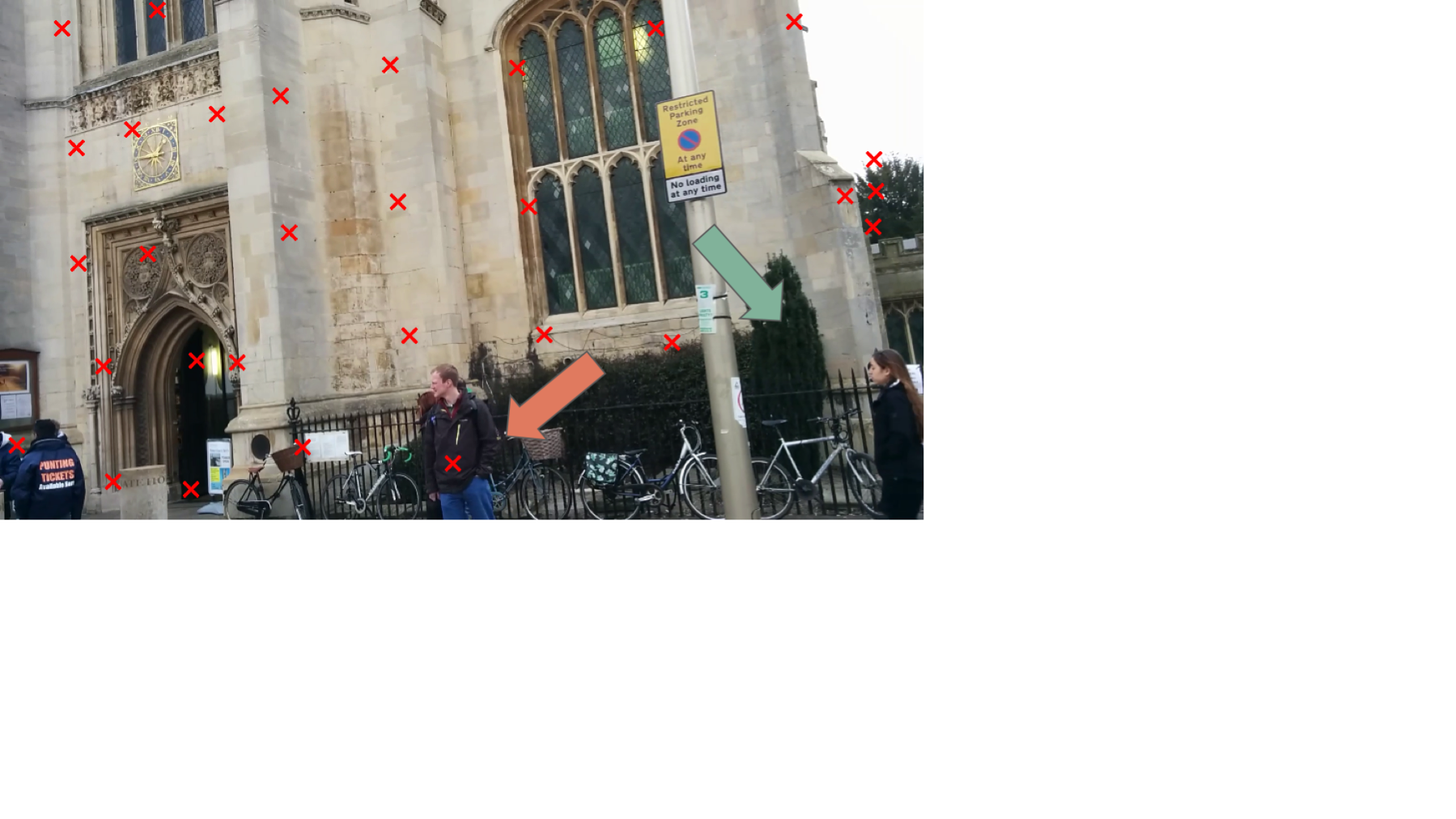}
        \caption{\small Scene-specific landmarks}
    \end{subfigure}
    }

    \centering
    \resizebox{0.9\linewidth}{!}{
        \begin{subfigure}[b]{0.49\linewidth}
            \includegraphics[width=\linewidth]{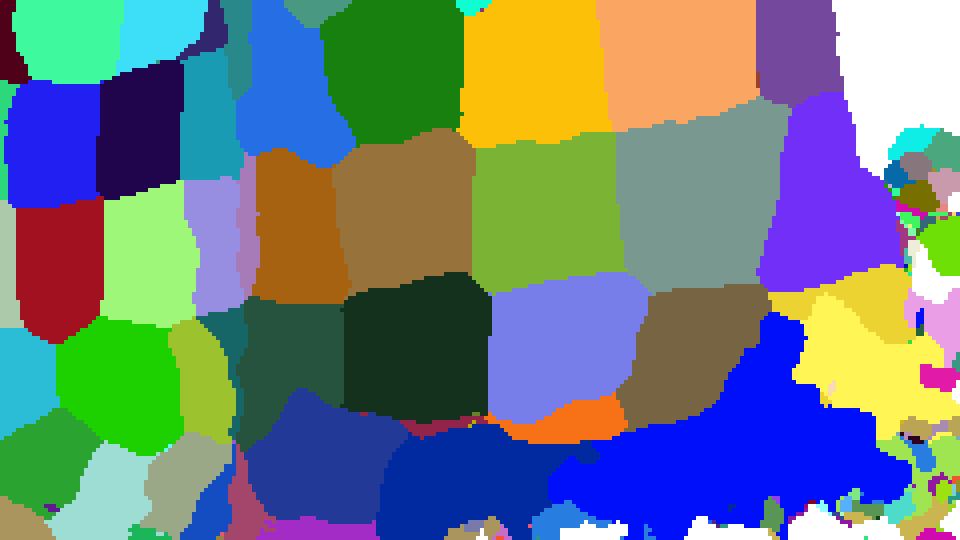}
            \caption{{\small Landmark segmentation map}}
        \end{subfigure}   
        \begin{subfigure}[b]{0.49\linewidth}
            \includegraphics[width=\linewidth]{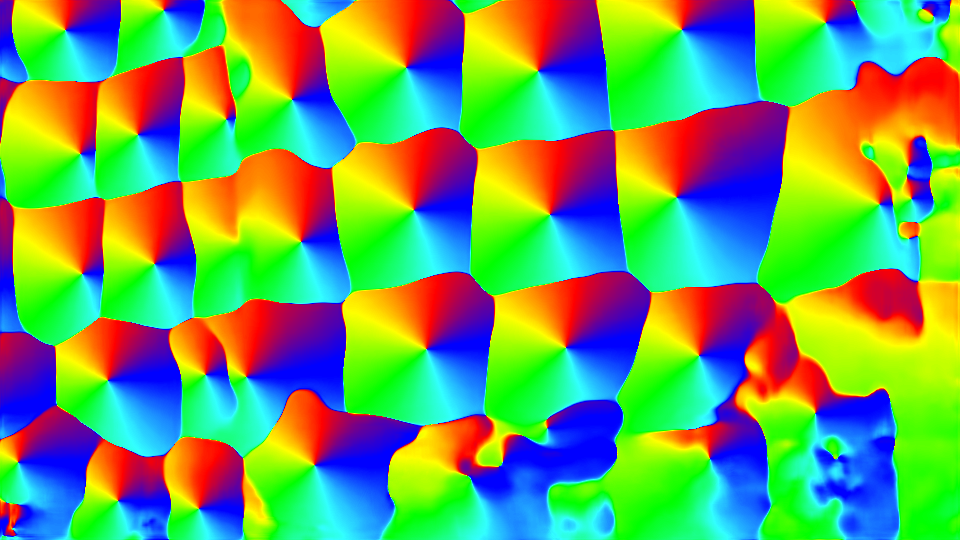}
            \caption{\small Landmark voting map}
        \end{subfigure}    
    }
    \caption{Comparison of (a) SIFT features and (b) the proposed scene-specific landmarks. (c-d) The  scene-specific landmarks in (b) are obtained based on (c) landmark segmentation map and (d) landmark location voting map.
    }
    \label{Fig:point detection}
\end{figure}

%% file: subsections/experiment/ComparisonTable.tex
\begin{table*}[tbh]
    \centering
    \resizebox{0.85\linewidth}{!}{
    \begin{tabular}{l|c|cc|ccc}
        \hline
    & \multicolumn{1}{c|}{SSL} & \multicolumn{2}{c|}{SfM} &   \multicolumn{3}{c}{Scene Coordinate}\\
    \hline
    \multicolumn{1}{c|}{} & VS-Net~(Ours) & AS~\cite{sattler2012improving} & HF-Net~\cite{sarlin2019coarse} & HSC-Net~\cite{li2020hierarchical} & Reg~\cite{li2020hierarchical} & DSAC++~\cite{brachmann2018learning} \\
    \hline
    Chess            &     $\mathbf{1.5}$cm, $\mathbf{0.5}^\circ$     &4cm, $1.96^\circ$      &  2.6cm, 0.9$^\circ$   & 2.1cm, $0.7^\circ$        &   2.1cm, $1.0^\circ$     &  $\mathbf{1.5}$cm, $\mathbf{0.5}^\circ$      \\
    Fire             &     $\mathbf{1.9}$cm, $\mathbf{0.8}^\circ$     & 3cm, $1.53^\circ$      &   2.7cm, 1.0$^\circ$  & 2.2cm, $0.9^\circ$        &    2.4cm, $0.9^\circ$    &  2.0cm, $0.9^\circ$      \\
    Heads            &     $\mathbf{1.2}$cm, $\mathbf{0.7}^\circ$     &  2cm, $1.45^\circ$      & 1.4cm, 0.9$^\circ$   & $\mathbf{1.2}$cm, $0.9^\circ$        &  $\mathbf{1.2}$cm, $0.8^\circ$      &  1.3cm, $0.8^\circ$      \\
    Office           &     $\mathbf{2.1}$cm, $\mathbf{0.6}^\circ$     & 9cm, $3.61^\circ$      &  4.3cm, 1.2$^\circ$    & 2.7cm, $0.8^\circ$        &  3.1cm, $0.9^\circ$      &  2.6cm, $0.7^\circ$       \\
    Pumpkin          &     $\mathbf{3.7}$cm, $\mathbf{1.0}^\circ$     &8cm, $3.10^\circ$      &  5.8cm, 1.6$^\circ$    & 4.0cm, $\mathbf{1.0}^\circ$        &    4.3cm, $1.1^\circ$    &   4.3cm, $1.1^\circ$      \\
    Kitchen          &     $\mathbf{3.6}$cm, $\mathbf{1.1}^\circ$     &7cm, $3.37^\circ$      &  5.3cm, 1.6$^\circ$    & 4.0cm, $1.8^\circ$        &   4.5cm, $1.4^\circ$     &  3.8cm, $\mathbf{1.1}^\circ$      \\
    Stairs           &     $\mathbf{2.8}$cm, $\mathbf{0.8}^\circ$     &3cm, $2.22^\circ$      &  7.2cm, 1.9$^\circ$    & 3.1cm, $\mathbf{0.8}^\circ$       &   3.8cm, $0.9^\circ$     &  9.1cm, $2.5^\circ$      \\
    \hline
    Avg           &    $\mathbf{2.4}$cm, $\mathbf{0.8}^\circ$        &  5.1cm, $2.5^\circ$ &  4.2cm, $1.3^\circ$     & 2.7cm, $1.0^\circ$    &  3.1cm, $1.0^\circ$  & 3.5cm, $1.1^\circ$ \\
    \hline
    \hline
    GreatCourt      &   $\mathbf{0.22}$m, $\mathbf{0.1}^\circ$       &-         &   0.76m, $0.3^\circ$   & 0.28m, $0.2^\circ$        &    1.25m, $0.6^\circ$   &   0.40m, $0.2^\circ$      \\
    KingsCollege    &     $\mathbf{0.16}$m, $\mathbf{0.2}^\circ$     &0.42m, $0.55^\circ$      &   0.34m, $0.4^\circ$   & 0.18m, $0.3^\circ$        &   0.21m, $0.3^\circ$    &   0.18m, $0.3^\circ$      \\
    OldHospital     &     $\mathbf{0.16}$m, $\mathbf{0.3}^\circ$     &0.44m, $1.01^\circ$      &  0.43m, $0.6^\circ$    & 0.19m, $\mathbf{0.3}^\circ$        &   0.21m, $\mathbf{0.3}^\circ$    &      0.20m, $\mathbf{0.3}^\circ$      \\
    ShopFacade      &     $\mathbf{0.06}$m, $\mathbf{0.3}^\circ$     &0.12m, $0.40^\circ$      &  0.09m, $0.4^\circ$    & $\mathbf{0.06}$m, $\mathbf{0.3}^\circ$        &   $\mathbf{0.06}$m, $\mathbf{0.3}^\circ$    &     $\mathbf{0.06}$m, $\mathbf{0.3}^\circ$     \\
    St.MarysChurch   &     $\mathbf{0.08}$m, $\mathbf{0.3}
    ^\circ$     &0.19m, $0.54^\circ$      &   0.16m, $0.5^\circ$   & 0.09m, $\mathbf{0.3}^\circ$        &   0.16m, $0.5^\circ$    &      0.13m, $0.4^\circ$      \\
        \hline 
    Avg           &    $\mathbf{0.136}$m, $\mathbf{0.24}^\circ$        &  - &  0.356m, $0.31^\circ$    & 0.160m, $0.28^\circ$    &  0.378m, $0.40^\circ$  & 0.194m, $0.3^\circ$ \\
    \hline
    \end{tabular}
    }
    \caption{Visual localization accuracy of state-of-the-art methods.
    We evaluate the localization performance by median positional error and angular error. 
    The bar~(-) means Active Search fails in the GreatCourt. 
    }
    \label{tab: camera localization}
\end{table*}

%% file: subsections/experiment/PoseAccuracyFigure/PoseAccuracyFigure.tex
\begin{figure*}[t]
    \centering
    \resizebox{0.9\linewidth}{!}{
        \begin{subfigure}{\linewidth}
         \includegraphics[width=\linewidth, trim={0mm 0 0mm 0}, clip]{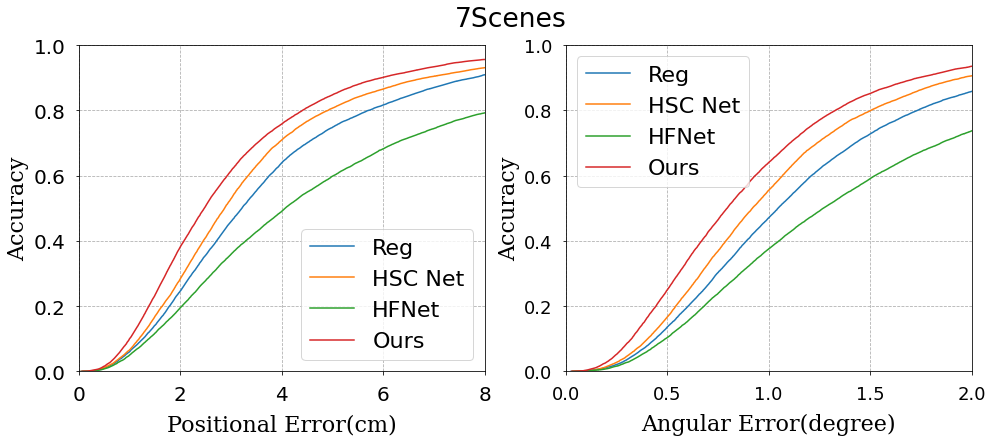}
        \end{subfigure}
        \begin{subfigure}{\linewidth}
        \includegraphics[width=\linewidth, trim={0mm 0 0mm 0}, clip]{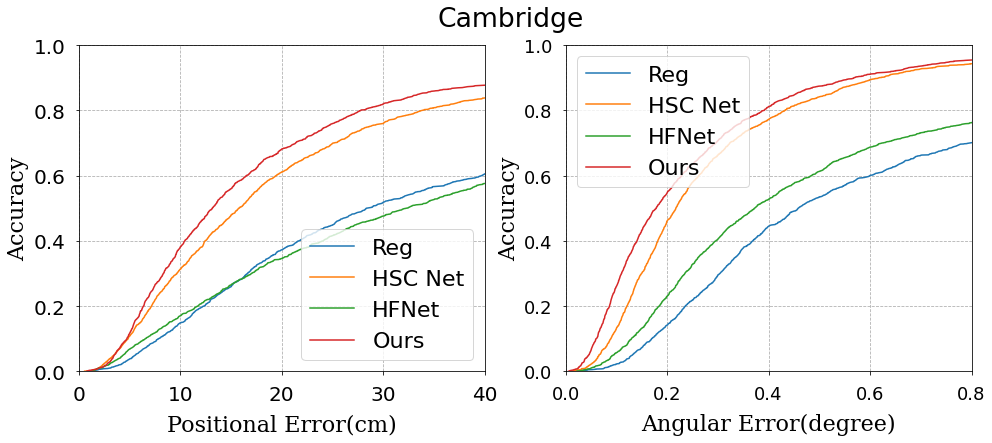}
        \end{subfigure}
    }
    \caption{Cumulative pose error distribution of representative methods. 
    For each dataset,
    we combine the poses of all scenes together, and count the ratio of poses under an increasing error threshold.
    }
    \label{Fig:cumulative result}
\end{figure*}

%% file: subsections/experiment/HardCase/HardCase.tex
\begin{figure}[t]
    \centering
    \includegraphics[width=1.0\linewidth, trim={76mm 90mm 92mm 10mm}, clip]{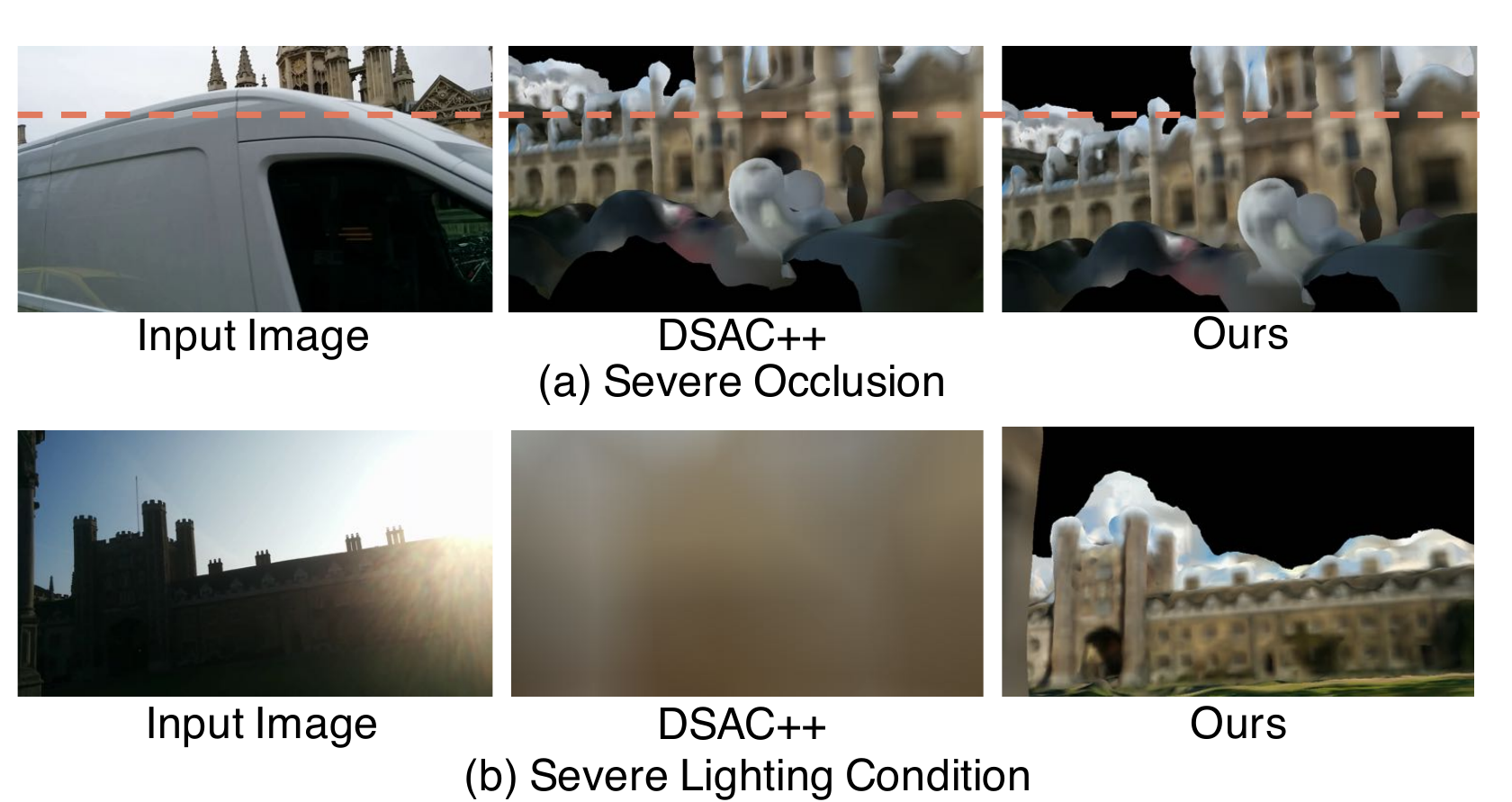}
    \caption{Localization in tough environments.
    We quantitatively compare VS-Net and DSAC++ by projecting the model into the original frames with estimated camera poses.
    }
    \vspace{-3mm}
    \label{Fig:quantitative comparison}
\end{figure}

%% file: subsections/experiment.tex
{In this section, we conduct a comparison with state-of-the-art methods and perform an ablation study to investigate individual components. Implementation details and extra results can be found in the supplementary materials.}


\subsection{Dataset}
We evaluate our VS-Net on two standard visual localization benchmark datasets.
(1) {\it Microsoft 7-Scenes Dataset} consists of seven static indoor scenes, which provides data recorded by a Kinect RGB-D sensor.
The 3D surfaces, along with the camera poses, are computed from KinectFusion~\cite{newcombe2011kinectfusion}.
(2) {\it Cambridge Landmarks Dataset} contains six urban scenes.
The images are collected by a smartphone and the camera poses are recovered from SfM.
The \textit{GreatCourt} and the \textit{King'sCollege} are two challenging scenes that are affected by varying lighting conditions and dynamic objects.
We reconstruct a dense 3D surface through multi-view stereo for each scene with given camera poses.

\subsection{Comparison with State-of-the-arts}

Previous visual localization systems that achieve good performance are SfM-based frameworks \cite{donoser2014discriminative,li2010location,zeisl2015camera,sarlin2019coarse} and scene coordinate regression frameworks \cite{li2020hierarchical, brachmann2018learning}.
We compare VS-Net with these two frameworks on the 7-Scenes dataset and  Cambridge Landmarks dataset.
We do not present the results of neural pose regression~\cite{kendall2015posenet,kendall2017geometric,brahmbhatt2018geometry, huang2019prior} because their pose accuracies are not competitive enough.
There are representative SfM-based visual localization methods \cite{sattler2012improving, sarlin2019coarse}. Active Search~\cite{sattler2012improving} utilizes SIFT features, which is the state-of-the-art method using handcrafted features with a priority-based matching algorithm.
HF-Net~\cite{sarlin2019coarse} computes camera poses with NetVLAD~\cite{arandjelovic2016netvlad} and SuperPoint~\cite{detone2018superpoint}, which are learned image features and local features, respectively.
For scene coordinate regression methods, we select Reg~\cite{li2020hierarchical}, DSAC++~\cite{brachmann2018learning}, and HSC-Net~\cite{li2020hierarchical} for comparison.
Reg~\cite{li2020hierarchical} is a regression-only method, which directly regresses the scene coordinates from a query image.
It is regarded as the baseline in scene coordinate regression methods.
DSAC++~\cite{brachmann2018learning} designs a pose hypothesis selection algorithm based on Reg.
HSC-Net~\cite{li2020hierarchical} is a state-of-the-art scene coordinate regression method, which predicts hierarchical scene coordinates to improve the localization performance.

We compare the localization accuracy of VS-Net with the above mentioned methods. Positional error and angular error are the main metrics for evaluating pose accuracies.
Table~\ref{tab: camera localization} presents the median of pose errors in each individual scene.
SfM-based methods, including Active Search and HF-Net, produce mediocre results because the general-purpose features are not accurate enough.
VS-Net achieves better performances in all scenes.
Even compared with improved scene coordinate regression methods (HSC-Net and DSAC++), VS-Net still outperforms them in most scenes.
Fig.~\ref{Fig:cumulative result} shows the cumulative distributions of overall pose error across scenes, which illustrate that the holistic performance of VS-Net is better than the others.
Moreover, VS-Net is able to obtain high-quality poses even though running into challenging cases (Fig.~\ref{Fig:quantitative comparison}) while DSAC++, an improved scene coordinate regression method, fails.

\subsection{Ablation study}

\noindent {\bf Scene-specific landmarks vs. scene coordinates.} We propose a new 2D-to-3D correspondence representation, the scene-specific landmarks, to replace the pixel-wise scene coordinates in deep learning based methods \cite{li2020hierarchical, brachmann2018learning}.
To compare these two representations, 
we remove the vote decoder in our VS-Net and directly regress pixel-wise scene coordinates with our segmentation decoder, which is similar to the Reg~\cite{li2020hierarchical}, and keep other settings the same.
Its average median errors of camera poses in the Microsoft 7-Scenes dataset and Cambridge Landmark dataset are 36.5cm/$16^\circ$ and 99cm/$1.7^\circ$, while our scene-specific landmarks achieves 2.4cm/$0.8^\circ$ and 14cm/$0.24^\circ$.
It is also far worse than Reg because the large receptive field of our VS-Net impacts scene coordinate regression.

\input{subsections/experiment/SegmentationComparisonTable}



\begin{table}[t]
    \centering
    \resizebox{1.0\linewidth}{!}{
    \begin{tabular}{c|c|c|c|c|c|c|c}
    \hline
    Size & 1.50m & $\mathbf{1.75}$m & 2.00m & 2.25m & 2.50m & 2.75m & 3.00m\\
    \hline 
    Num. & 7418 & $\mathbf{5333}$ & 4089 & 3278 & 2603 & 2099 & 1804\\
    Pos. & 16cm & $\mathbf{15}$cm & 16cm & 15cm & 17cm & 17cm & 18cm\\
    Ang. & $0.3^\circ$ & $\mathbf{0.2}^\circ$ & $0.2^\circ$ & $0.3^\circ$ & $0.3^\circ$ & $0.3^\circ$ & $0.3^\circ$\\
    \hline
    \hline
    Size & 10.0cm & 12.5cm & $\mathbf{15.0}$cm & 17.5cm & 20.0cm & 25.0cm & 30.0cm\\
    \hline 
    Num. & 10918 & 6501 & $\mathbf{4330}$ & 3013 & 2280 & 1409 & 925\\
    Pos. & 1.54cm & 1.52cm & $\mathbf{1.52}$cm & 1.52cm & 1.57cm & 1.58cm & 1.65cm\\
    Ang. & $0.54^\circ$ & $0.53^\circ$ & $\mathbf{0.50}^\circ$ & $0.50^\circ$ & $0.54^\circ$ & $0.54^\circ$ & $0.55^\circ$\\
    \hline
    \end{tabular}
    }
    \caption{Localization accuracy in \textit{King's College} (above) and \textit{chess} (below) with different patch sizes.
    }
    \label{tab: segmentation size ablation}
\end{table}

\begin{figure}[t]
    \centering
    \resizebox{\linewidth}{!}{
    \begin{subfigure}[b]{0.3\linewidth}
        \includegraphics[width=\linewidth]{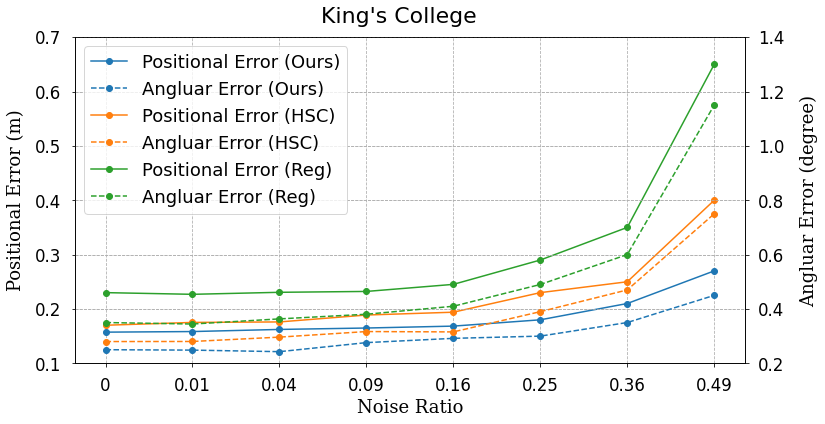}   
    \end{subfigure}
    }
    \caption{Localization pose errors of different methods when a noisy occlusion block of different sizes (noise ratios) is introduced into query images.
    }
    \label{Fig: noise ratio}
\end{figure}

\noindent {\bf Localization accuracy with different patch sizes.} 
Our landmarks are generated through 3D surface over-segmentation, where patch size is a hyper-parameter that determines the number and distribution of landmarks in VS-Net.
To explore the influence of different patch sizes, we train and evaluate VS-Net with the landmark segmentation map and pixel-wise voting map predicted with different patch sizes.
We show the corresponding landmark numbers and median pose errors of different patch sizes in Table \ref{tab: segmentation size ablation}.
The best choice of patch size for \textit{King's College} and \textit{chess} are 1.75m and 15cm, respectively.
Both a smaller size and a larger size would deteriorate the localization accuracy but do not severely impact it, which indicates that VS-Net is not very sensitive to the patch size.

\noindent {\bf Localization with challenging occlusions.} 
To evaluate the robustness of visual localization systems against environmental noise, we use a noisy occlusion block that contains 3-channel random noise ranging from 0 to 255 to cover a portion of the query image randomly.
The height and width of the noise blocks are set as 10\%-70\% of the image, which correspond to 1\%-49\% noise ratio and can indicate different levels of noise interference.
As shown in Fig.~\ref{Fig: noise ratio}, we compare the median pose error of our VS-Net with those of HSC-Net~\cite{li2020hierarchical} and Reg~\cite{li2020hierarchical} under different noise ratios.
VS-Net consistently results in lower pose errors.

\noindent {\bf Landmark segmentation with prototype-based triplet loss.}
To address the problem of the too large number of landmarks in our landmark segmentation sub-task, we propose the prototype-based triplet loss for our VS-Net.
We conduct an experiment in Table~\ref{tab: segmentation} to compare segmentation loss functions on the 7-Scenes dataset.
The conventional cross entropy loss does not work for VS-Net because it consumes much computation and memory. 
We present its theoretical computational and memory costs for a single image of size $640\times480$ in the braces.
We also test using the prototype-based triplet loss for landmark segmentation but without the $k$NN negative mining, which is able to train VS-Net but achieves inferior performances.
Our complete prototype-based triplet has low latency and computation complexity while maintaining superior performance.

%% file: subsections/experiment/SegmentationComparisonTable.tex
\begin{table}[t]
    \centering
    \resizebox{1.0\linewidth}{!}{
    \begin{tabular}{l|c|ccc}
        \hline
    & Cross Entropy & \multicolumn{3}{c}{Proto. Triplet (Avg FLOPS/Bytes)} \\
    & (FLOPS/Bytes) & A.L. & w/o $k$NN &  w/ $k$NN\\
    \hline
    \hline
    Chess            &     n/a (30G/5G) & 470   & 0.45 (0.01G/1M)      &  0.80 (0.06G/9M)   \\
    Fire             &     n/a (35G/6G)  & 662  & 0.34 (0.01G/1M)     &  0.69 (0.10G/14M)      \\
    Heads            &     n/a (69G/12G) & 826    &  0.25 (0.01G/1M)     & 0.58 (0.21G/32M)     \\
    Office           &     n/a (34G/6G) & 409   & 0.4 (0.01G/1M)     &  0.75 (0.06G/9M)  \\
    Pumpkin          &     n/a (32G/5G) & 519   & 0.43 (0.01G/1M)     &  0.61 (0.07G/10M)    \\
    Kitchen          &     n/a (44G/7G) & 496   & 0.26 (0.01G/1M)    &  0.58 (0.09G/13M)       \\
    Stairs           &     n/a (95G/16G) & 224    & 0.45 (0.01G/1M)     &  0.68 (0.09G/13M)      \\
    \hline
    \end{tabular}
    }
    \caption{Segmentation accuracy on \textit{7-Scenes} by our VS-Net with different segmentation losses. 
    n/a denotes that cross entropy loss alone already occupies too much memory even for a single image and cannot be used in practice.
    A.L. denotes the average active labels of images in the scene.
    }
    \label{tab: segmentation}
\end{table}

%% file: subsections/conclusion.tex

In this paper, we have proposed a novel visual localization framework that represents the map by patches and landmarks, and design a neural network VS-Net to detect the scene-specific landmarks on images. The experiments on the public datasets demonstrate the effectiveness of the proposed framework.
Utilizing hierarchical spatial structure and temporal information has been proved beneficial in both SfM-based methods and scene coordinate regression.
Exploring how to improve scene-specific landmarks with these strategies will be the direction of our future work.

%% file: cvpr.bbl
\begin{thebibliography}{10}\itemsep=-1pt

\bibitem{agarwal2011building}
Sameer Agarwal, Yasutaka Furukawa, Noah Snavely, Ian Simon, Brian Curless,
  Steven~M Seitz, and Richard Szeliski.
\newblock Building rome in a day.
\newblock {\em Communications of the ACM}, 54(10):105--112, 2011.

\bibitem{antonio1992faster}
Franklin Antonio.
\newblock Faster line segment intersection.
\newblock In {\em Graphics Gems III (IBM Version)}, pages 199--202. Elsevier,
  1992.

\bibitem{arandjelovic2016netvlad}
Relja Arandjelovic, Petr Gronat, Akihiko Torii, Tomas Pajdla, and Josef Sivic.
\newblock Netvlad: Cnn architecture for weakly supervised place recognition.
\newblock In {\em Proceedings of the IEEE conference on computer vision and
  pattern recognition}, pages 5297--5307, 2016.

\bibitem{arth2009wide}
Clemens Arth, Daniel Wagner, Manfred Klopschitz, Arnold Irschara, and Dieter
  Schmalstieg.
\newblock Wide area localization on mobile phones.
\newblock In {\em 2009 8th ieee international symposium on mixed and augmented
  reality}, pages 73--82. IEEE, 2009.

\bibitem{aziere2019ensemble}
Nicolas Aziere and Sinisa Todorovic.
\newblock Ensemble deep manifold similarity learning using hard proxies.
\newblock In {\em Proceedings of the IEEE Conference on Computer Vision and
  Pattern Recognition}, pages 7299--7307, 2019.

\bibitem{bay2006surf}
Herbert Bay, Tinne Tuytelaars, and Luc Van~Gool.
\newblock {SURF}: Speeded up robust features.
\newblock In {\em Proceedings of the European conference on computer vision},
  pages 404--417. Springer, 2006.

\bibitem{brachmann2017dsac}
Eric Brachmann, Alexander Krull, Sebastian Nowozin, Jamie Shotton, Frank
  Michel, Stefan Gumhold, and Carsten Rother.
\newblock Dsac-differentiable ransac for camera localization.
\newblock In {\em Proceedings of the IEEE Conference on Computer Vision and
  Pattern Recognition}, pages 6684--6692, 2017.

\bibitem{brachmann2018learning}
Eric Brachmann and Carsten Rother.
\newblock Learning less is more-6d camera localization via 3d surface
  regression.
\newblock In {\em Proceedings of the IEEE Conference on Computer Vision and
  Pattern Recognition}, pages 4654--4662, 2018.

\bibitem{brachmann2019expert}
Eric Brachmann and Carsten Rother.
\newblock Expert sample consensus applied to camera re-localization.
\newblock In {\em Proceedings of the IEEE International Conference on Computer
  Vision}, pages 7525--7534, 2019.

\bibitem{brahmbhatt2018geometry}
Samarth Brahmbhatt, Jinwei Gu, Kihwan Kim, James Hays, and Jan Kautz.
\newblock Geometry-aware learning of maps for camera localization.
\newblock In {\em Proceedings of the IEEE Conference on Computer Vision and
  Pattern Recognition}, pages 2616--2625, 2018.

\bibitem{budvytis2019large}
Ignas Budvytis, Marvin Teichmann, Tomas Vojir, and Roberto Cipolla.
\newblock Large scale joint semantic re-localisation and scene understanding
  via globally unique instance coordinate regression.
\newblock {\em arXiv preprint arXiv:1909.10239}, 2019.

\bibitem{camposeco2019hybrid}
Federico Camposeco, Andrea Cohen, Marc Pollefeys, and Torsten Sattler.
\newblock Hybrid scene compression for visual localization.
\newblock In {\em Proceedings of the IEEE Conference on Computer Vision and
  Pattern Recognition}, pages 7653--7662, 2019.

\bibitem{DBLP:journals/corr/ChenPSA17}
Liang{-}Chieh Chen, George Papandreou, Florian Schroff, and Hartwig Adam.
\newblock Rethinking atrous convolution for semantic image segmentation.
\newblock {\em CoRR}, abs/1706.05587, 2017.

\bibitem{chen2017deeplab}
Liang-Chieh Chen, George Papandreou, Iasonas Kokkinos, Kevin Murphy, and Alan~L
  Yuille.
\newblock Deeplab: Semantic image segmentation with deep convolutional nets,
  atrous convolution, and fully connected crfs.
\newblock {\em IEEE transactions on pattern analysis and machine intelligence},
  40(4):834--848, 2017.

\bibitem{detone2018superpoint}
Daniel DeTone, Tomasz Malisiewicz, and Andrew Rabinovich.
\newblock Superpoint: Self-supervised interest point detection and description.
\newblock In {\em Proceedings of the IEEE Conference on Computer Vision and
  Pattern Recognition Workshops}, pages 224--236, 2018.

\bibitem{donoser2014discriminative}
Michael Donoser and Dieter Schmalstieg.
\newblock Discriminative feature-to-point matching in image-based localization.
\newblock In {\em Proceedings of the IEEE Conference on Computer Vision and
  Pattern Recognition}, pages 516--523, 2014.

\bibitem{DBLP:conf/cvpr/DusmanuRPPSTS19}
Mihai Dusmanu, Ignacio Rocco, Tom{\'{a}}s Pajdla, Marc Pollefeys, Josef Sivic,
  Akihiko Torii, and Torsten Sattler.
\newblock D2-net: {A} trainable {CNN} for joint description and detection of
  local features.
\newblock In {\em {IEEE} Conference on Computer Vision and Pattern Recognition,
  {CVPR} 2019, Long Beach, CA, USA, June 16-20, 2019}, pages 8092--8101, 2019.

\bibitem{fischler1981random}
Martin~A Fischler and Robert~C Bolles.
\newblock Random sample consensus: a paradigm for model fitting with
  applications to image analysis and automated cartography.
\newblock {\em Communications of the ACM}, 24(6):381--395, 1981.

\bibitem{ge2020self}
Yixiao Ge, Haibo Wang, Feng Zhu, Rui Zhao, and Hongsheng Li.
\newblock Self-supervising fine-grained region similarities for large-scale
  image localization.
\newblock {\em arXiv preprint arXiv:2006.03926}, 2020.

\bibitem{he2020pvn3d}
Yisheng He, Wei Sun, Haibin Huang, Jianran Liu, Haoqiang Fan, and Jian Sun.
\newblock Pvn3d: A deep point-wise 3d keypoints voting network for 6dof pose
  estimation.
\newblock In {\em Proceedings of the IEEE/CVF Conference on Computer Vision and
  Pattern Recognition}, pages 11632--11641, 2020.

\bibitem{huang2019prior}
Zhaoyang Huang, Yan Xu, Jianping Shi, Xiaowei Zhou, Hujun Bao, and Guofeng
  Zhang.
\newblock Prior guided dropout for robust visual localization in dynamic
  environments.
\newblock In {\em Proceedings of the IEEE International Conference on Computer
  Vision}, pages 2791--2800, 2019.

\bibitem{imperoli2016active}
Marco Imperoli and Alberto Pretto.
\newblock Active detection and localization of textureless objects in cluttered
  environments.
\newblock {\em arXiv preprint arXiv:1603.07022}, 2016.

\bibitem{kendall2017geometric}
Alex Kendall and Roberto Cipolla.
\newblock Geometric loss functions for camera pose regression with deep
  learning.
\newblock In {\em Proceedings of the IEEE Conference on Computer Vision and
  Pattern Recognition}, pages 5974--5983, 2017.

\bibitem{kendall2015posenet}
Alex Kendall, Matthew Grimes, and Roberto Cipolla.
\newblock Posenet: A convolutional network for real-time 6-dof camera
  relocalization.
\newblock In {\em Proceedings of the IEEE international conference on computer
  vision}, pages 2938--2946, 2015.

\bibitem{li2020hierarchical}
Xiaotian Li, Shuzhe Wang, Yi Zhao, Jakob Verbeek, and Juho Kannala.
\newblock Hierarchical scene coordinate classification and regression for
  visual localization.
\newblock In {\em Proceedings of the IEEE/CVF Conference on Computer Vision and
  Pattern Recognition}, pages 11983--11992, 2020.

\bibitem{li2010location}
Yunpeng Li, Noah Snavely, and Daniel~P Huttenlocher.
\newblock Location recognition using prioritized feature matching.
\newblock In {\em European conference on computer vision}, pages 791--804.
  Springer, 2010.

\bibitem{lin2020unsupervised}
Yutian Lin, Lingxi Xie, Yu Wu, Chenggang Yan, and Qi Tian.
\newblock Unsupervised person re-identification via softened similarity
  learning.
\newblock In {\em Proceedings of the IEEE/CVF Conference on Computer Vision and
  Pattern Recognition}, pages 3390--3399, 2020.

\bibitem{liu2019gift}
Yuan Liu, Zehong Shen, Zhixuan Lin, Sida Peng, Hujun Bao, and Xiaowei Zhou.
\newblock Gift: Learning transformation-invariant dense visual descriptors via
  group cnns.
\newblock In {\em Advances in Neural Information Processing Systems}, pages
  6990--7001, 2019.

\bibitem{long2015fully}
Jonathan Long, Evan Shelhamer, and Trevor Darrell.
\newblock Fully convolutional networks for semantic segmentation.
\newblock In {\em Proceedings of the IEEE conference on computer vision and
  pattern recognition}, pages 3431--3440, 2015.

\bibitem{lowe2004distinctive}
David~G Lowe.
\newblock Distinctive image features from scale-invariant keypoints.
\newblock {\em International journal of computer vision}, 60(2):91--110, 2004.

\bibitem{morel2009asift}
Jean-Michel Morel and Guoshen Yu.
\newblock Asift: A new framework for fully affine invariant image comparison.
\newblock {\em SIAM journal on imaging sciences}, 2(2):438--469, 2009.

\bibitem{movshovitz2017no}
Yair Movshovitz-Attias, Alexander Toshev, Thomas~K Leung, Sergey Ioffe, and
  Saurabh Singh.
\newblock No fuss distance metric learning using proxies.
\newblock In {\em Proceedings of the IEEE International Conference on Computer
  Vision}, pages 360--368, 2017.

\bibitem{newcombe2011kinectfusion}
Richard~A Newcombe, Shahram Izadi, Otmar Hilliges, David Molyneaux, David Kim,
  Andrew~J Davison, Pushmeet Kohli, Jamie Shotton, Steve Hodges, and Andrew~W
  Fitzgibbon.
\newblock Kinectfusion: Real-time dense surface mapping and tracking.
\newblock In {\em ISMAR}, volume~11, pages 127--136, 2011.

\bibitem{oberweger2018making}
Markus Oberweger, Mahdi Rad, and Vincent Lepetit.
\newblock Making deep heatmaps robust to partial occlusions for 3d object pose
  estimation.
\newblock In {\em Proceedings of the European Conference on Computer Vision
  (ECCV)}, pages 119--134, 2018.

\bibitem{ono2018lf}
Yuki Ono, Eduard Trulls, Pascal Fua, and Kwang~Moo Yi.
\newblock Lf-net: learning local features from images.
\newblock In {\em Advances in neural information processing systems}, pages
  6234--6244, 2018.

\bibitem{Papon13CVPR}
Jeremie Papon, Alexey Abramov, Markus Schoeler, and Florentin
  W\"{o}rg\"{o}tter.
\newblock Voxel cloud connectivity segmentation - supervoxels for point clouds.
\newblock In {\em Computer Vision and Pattern Recognition (CVPR), 2013 IEEE
  Conference on}, Portland, Oregon, June 22-27 2013.

\bibitem{pavlakos20176}
Georgios Pavlakos, Xiaowei Zhou, Aaron Chan, Konstantinos~G Derpanis, and
  Kostas Daniilidis.
\newblock 6-dof object pose from semantic keypoints.
\newblock In {\em 2017 IEEE international conference on robotics and automation
  (ICRA)}, pages 2011--2018. IEEE, 2017.

\bibitem{peng2019pvnet}
Sida Peng, Yuan Liu, Qixing Huang, Xiaowei Zhou, and Hujun Bao.
\newblock Pvnet: Pixel-wise voting network for 6dof pose estimation.
\newblock In {\em Proceedings of the IEEE Conference on Computer Vision and
  Pattern Recognition}, pages 4561--4570, 2019.

\bibitem{qian2019softtriple}
Qi Qian, Lei Shang, Baigui Sun, Juhua Hu, Hao Li, and Rong Jin.
\newblock Softtriple loss: Deep metric learning without triplet sampling.
\newblock In {\em Proceedings of the IEEE International Conference on Computer
  Vision}, pages 6450--6458, 2019.

\bibitem{qin2018vins}
Tong Qin, Peiliang Li, and Shaojie Shen.
\newblock Vins-mono: A robust and versatile monocular visual-inertial state
  estimator.
\newblock {\em IEEE Transactions on Robotics}, 34(4):1004--1020, 2018.

\bibitem{revaud2019r2d2}
Jerome Revaud, Cesar De~Souza, Martin Humenberger, and Philippe Weinzaepfel.
\newblock R2d2: Reliable and repeatable detector and descriptor.
\newblock In {\em Advances in Neural Information Processing Systems}, pages
  12405--12415, 2019.

\bibitem{ronneberger2015u}
Olaf Ronneberger, Philipp Fischer, and Thomas Brox.
\newblock U-net: Convolutional networks for biomedical image segmentation.
\newblock In {\em International Conference on Medical image computing and
  computer-assisted intervention}, pages 234--241. Springer, 2015.

\bibitem{rublee2011orb}
Ethan Rublee, Vincent Rabaud, Kurt Konolige, and Gary Bradski.
\newblock {ORB: An efficient alternative to SIFT or SURF}.
\newblock In {\em Proceedings of the IEEE international conference on Computer
  Vision (ICCV)}, pages 2564--2571. IEEE, 2011.

\bibitem{sarlin2019coarse}
Paul-Edouard Sarlin, Cesar Cadena, Roland Siegwart, and Marcin Dymczyk.
\newblock From coarse to fine: Robust hierarchical localization at large scale.
\newblock In {\em Proceedings of the IEEE Conference on Computer Vision and
  Pattern Recognition}, pages 12716--12725, 2019.

\bibitem{sattler2012improving}
Torsten Sattler, Bastian Leibe, and Leif Kobbelt.
\newblock Improving image-based localization by active correspondence search.
\newblock In {\em European conference on computer vision}, pages 752--765.
  Springer, 2012.

\bibitem{schonberger2016structure}
Johannes~L Schonberger and Jan-Michael Frahm.
\newblock Structure-from-motion revisited.
\newblock In {\em Proceedings of the IEEE Conference on Computer Vision and
  Pattern Recognition}, pages 4104--4113, 2016.

\bibitem{schoenberger2016mvs}
Johannes~Lutz Sch\"{o}nberger, Enliang Zheng, Marc Pollefeys, and Jan-Michael
  Frahm.
\newblock Pixelwise view selection for unstructured multi-view stereo.
\newblock In {\em European Conference on Computer Vision (ECCV)}, 2016.

\bibitem{schroff2015facenet}
Florian Schroff, Dmitry Kalenichenko, and James Philbin.
\newblock Facenet: A unified embedding for face recognition and clustering.
\newblock In {\em Proceedings of the IEEE conference on computer vision and
  pattern recognition}, pages 815--823, 2015.

\bibitem{shotton2013scene}
Jamie Shotton, Ben Glocker, Christopher Zach, Shahram Izadi, Antonio Criminisi,
  and Andrew Fitzgibbon.
\newblock Scene coordinate regression forests for camera relocalization in
  rgb-d images.
\newblock In {\em Proceedings of the IEEE Conference on Computer Vision and
  Pattern Recognition}, pages 2930--2937, 2013.

\bibitem{song2020hybridpose}
Chen Song, Jiaru Song, and Qixing Huang.
\newblock Hybridpose: 6d object pose estimation under hybrid representations.
\newblock In {\em Proceedings of the IEEE/CVF Conference on Computer Vision and
  Pattern Recognition}, pages 431--440, 2020.

\bibitem{valentin2015exploiting}
Julien Valentin, Matthias Nie{\ss}ner, Jamie Shotton, Andrew Fitzgibbon,
  Shahram Izadi, and Philip~HS Torr.
\newblock Exploiting uncertainty in regression forests for accurate camera
  relocalization.
\newblock In {\em Proceedings of the IEEE Conference on Computer Vision and
  Pattern Recognition}, pages 4400--4408, 2015.

\bibitem{wang2020atloc}
Bing Wang, Changhao Chen, Chris~Xiaoxuan Lu, Peijun Zhao, Niki Trigoni, and
  Andrew Markham.
\newblock Atloc: Attention guided camera localization.
\newblock In {\em Proceedings of the AAAI Conference on Artificial
  Intelligence}, volume~34, pages 10393--10401, 2020.

\bibitem{wang2020learning}
Qianqian Wang, Xiaowei Zhou, Bharath Hariharan, and Noah Snavely.
\newblock Learning feature descriptors using camera pose supervision.
\newblock {\em arXiv preprint arXiv:2004.13324}, 2020.

\bibitem{weinzaepfel2019visual}
Philippe Weinzaepfel, Gabriela Csurka, Yohann Cabon, and Martin Humenberger.
\newblock Visual localization by learning objects-of-interest dense match
  regression.
\newblock In {\em Proceedings of the IEEE/CVF Conference on Computer Vision and
  Pattern Recognition}, pages 5634--5643, 2019.

\bibitem{wu2011visualsfm}
Changchang Wu et~al.
\newblock Visualsfm: A visual structure from motion system.
\newblock 2011.

\bibitem{wu2017sampling}
Chao-Yuan Wu, R Manmatha, Alexander~J Smola, and Philipp Krahenbuhl.
\newblock Sampling matters in deep embedding learning.
\newblock In {\em Proceedings of the IEEE International Conference on Computer
  Vision}, pages 2840--2848, 2017.

\bibitem{xiao2017joint}
Tong Xiao, Shuang Li, Bochao Wang, Liang Lin, and Xiaogang Wang.
\newblock Joint detection and identification feature learning for person
  search.
\newblock In {\em Proceedings of the IEEE Conference on Computer Vision and
  Pattern Recognition}, pages 3415--3424, 2017.

\bibitem{xu2020selfvoxelo}
Yan Xu, Zhaoyang Huang, Kwan-Yee Lin, Xinge Zhu, Jianping Shi, Hujun Bao,
  Guofeng Zhang, and Hongsheng Li.
\newblock Selfvoxelo: Self-supervised lidar odometry with voxel-based deep
  neural networks.
\newblock {\em Conference on Robot Learning}, 2020.

\bibitem{xue2020learning}
Fei Xue, Xin Wu, Shaojun Cai, and Junqiu Wang.
\newblock Learning multi-view camera relocalization with graph neural networks.
\newblock In {\em 2020 IEEE/CVF Conference on Computer Vision and Pattern
  Recognition (CVPR)}, pages 11372--11381. IEEE, 2020.

\bibitem{DBLP:journals/corr/YuK15}
Fisher Yu and Vladlen Koltun.
\newblock Multi-scale context aggregation by dilated convolutions.
\newblock In {\em 4th International Conference on Learning Representations,
  {ICLR} 2016, San Juan, Puerto Rico, May 2-4, 2016, Conference Track
  Proceedings}, 2016.

\bibitem{zeisl2015camera}
Bernhard Zeisl, Torsten Sattler, and Marc Pollefeys.
\newblock Camera pose voting for large-scale image-based localization.
\newblock In {\em Proceedings of the IEEE International Conference on Computer
  Vision}, pages 2704--2712, 2015.

\bibitem{zhang2010efficient}
Guofeng Zhang, Zilong Dong, Jiaya Jia, Tien-Tsin Wong, and Hujun Bao.
\newblock Efficient non-consecutive feature tracking for structure-from-motion.
\newblock In {\em European Conference on Computer Vision}, pages 422--435.
  Springer, 2010.

\bibitem{zheng2019pose}
Liang Zheng, Yujia Huang, Huchuan Lu, and Yi Yang.
\newblock Pose-invariant embedding for deep person re-identification.
\newblock {\em IEEE Transactions on Image Processing}, 28(9):4500--4509, 2019.

\bibitem{zhong2020squeeze}
Zilong Zhong, Zhong~Qiu Lin, Rene Bidart, Xiaodan Hu, Ibrahim~Ben Daya, Zhifeng
  Li, Wei-Shi Zheng, Jonathan Li, and Alexander Wong.
\newblock Squeeze-and-attention networks for semantic segmentation.
\newblock In {\em Proceedings of the IEEE/CVF Conference on Computer Vision and
  Pattern Recognition}, pages 13065--13074, 2020.

\bibitem{zhong2020learning}
Zhun Zhong, Liang Zheng, Zhiming Luo, Shaozi Li, and Yi Yang.
\newblock Learning to adapt invariance in memory for person re-identification.
\newblock {\em IEEE Transactions on Pattern Analysis and Machine Intelligence},
  2020.

\bibitem{zhou2020kfnet}
Lei Zhou, Zixin Luo, Tianwei Shen, Jiahui Zhang, Mingmin Zhen, Yao Yao, Tian
  Fang, and Long Quan.
\newblock Kfnet: Learning temporal camera relocalization using kalman
  filtering.
\newblock In {\em Proceedings of the IEEE/CVF Conference on Computer Vision and
  Pattern Recognition}, pages 4919--4928, 2020.

\bibitem{zhu2017parallel}
Siyu Zhu, Tianwei Shen, Lei Zhou, Runze Zhang, Jinglu Wang, Tian Fang, and Long
  Quan.
\newblock Parallel structure from motion from local increment to global
  averaging.
\newblock {\em arXiv preprint arXiv:1702.08601}, 2017.

\end{thebibliography}
